\pdfoutput=1

\documentclass[11pt]{article}

\usepackage[final]{acl}

\usepackage{times}
\usepackage{latexsym}
\usepackage{enumitem}
\usepackage{amsmath}
\usepackage[T1]{fontenc}

\usepackage[utf8]{inputenc}

\usepackage{listings}
\usepackage{xcolor}  

\usepackage{microtype}

\usepackage{inconsolata}

\usepackage{graphicx}
\usepackage{tcolorbox}
\usepackage{etoolbox}
\usepackage{caption}
\usepackage{booktabs}
\usepackage{subcaption}
\usepackage{float}
\usepackage{amsfonts}
\AtBeginEnvironment{tcolorbox}{\small}

\title{How Can Input Reformulation Improve Tool Usage Accuracy in a Complex Dynamic Environment? A Study on $\tau$-bench}

\lstdefinestyle{dialogue}{
    basicstyle=\ttfamily\small,
    breaklines=true,
    showstringspaces=false,
    columns=fullflexible
}

\author{Venkatesh Mishra$^{\dagger*}$ \quad Amir Saeidi$^\dagger$\thanks{\ Equal Contribution} \quad Satyam Raj$^\dagger$ \quad Mutsumi Nakamura$^\dagger$  \\ \textbf{Jayanth Srinivasa}$^\ddagger$\quad  \textbf{Gaowen Liu}$^\ddagger$ \quad \textbf{Ali Payani}$^\ddagger$ \quad \textbf{Chitta Baral}$^\dagger$ \\\\ 
$^\dagger$Arizona State University \quad $^\ddagger$Cisco Research\\
\small{\texttt{\{vmishr23, ssaeidi1, chitta\}@asu.edu,}}
\small{\texttt{\{jasriniv, gaoliu, apayani\}@cisco.com}}
}

\begin{document}
\maketitle
\begin{abstract}
Recent advances in reasoning and planning capabilities of large language models (LLMs) have enabled their potential as autonomous agents capable of tool use in dynamic environments. However, in multi-turn conversational environments like $\tau$‑bench, these agents often struggle with consistent reasoning, adherence to domain-specific policies, and extracting correct information over a long horizon of tool-calls and conversation. To capture and mitigate these failures, we conduct a comprehensive manual analysis of the common errors occurring in the conversation trajectories. We then experiment with reformulations of inputs to the tool-calling agent for improvement in agent decision-making. Finally, we propose the \textbf{Input-Reformulation Multi-Agent (IRMA)} framework, which automatically reformulates user queries augmented with relevant domain rules and tool suggestions for the tool-calling agent to focus on. The results show that IRMA significantly outperforms ReAct, Function Calling, and Self-Reflection by \textbf{16.1\%}, \textbf{12.7\%}, and \textbf{19.1\%}, respectively, in overall pass\textasciicircum 5 scores. These findings highlight the superior reliability and consistency of IRMA compared to other methods in dynamic environments.


\end{abstract}



\section{Introduction}
\label{sec:introduction}

\begin{figure}[ht]
    \centering     \includegraphics[width=0.95\linewidth]{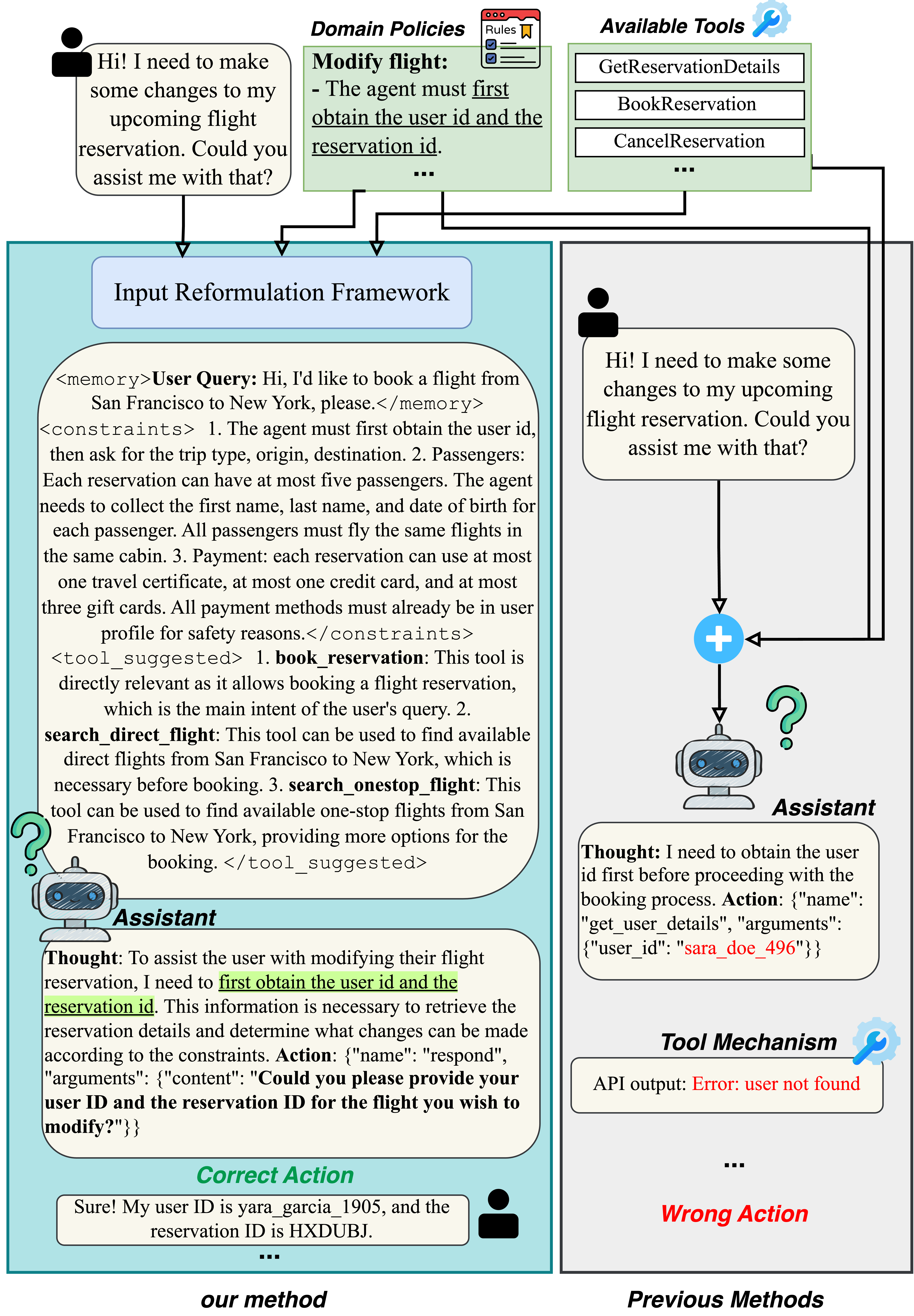}
    \caption{Comparison of the IRMA framework with other methods. The Input Reformulation framework takes the user query, domain policies, and available tools to generate a structured input for the assistant agent. This augmented input enables the assistant to make more accurate decisions from the first turn.}
    \label{fig:first_teaser}
\end{figure}

\begin{figure*}[t]
    \centering     \includegraphics[width=1.0\linewidth]{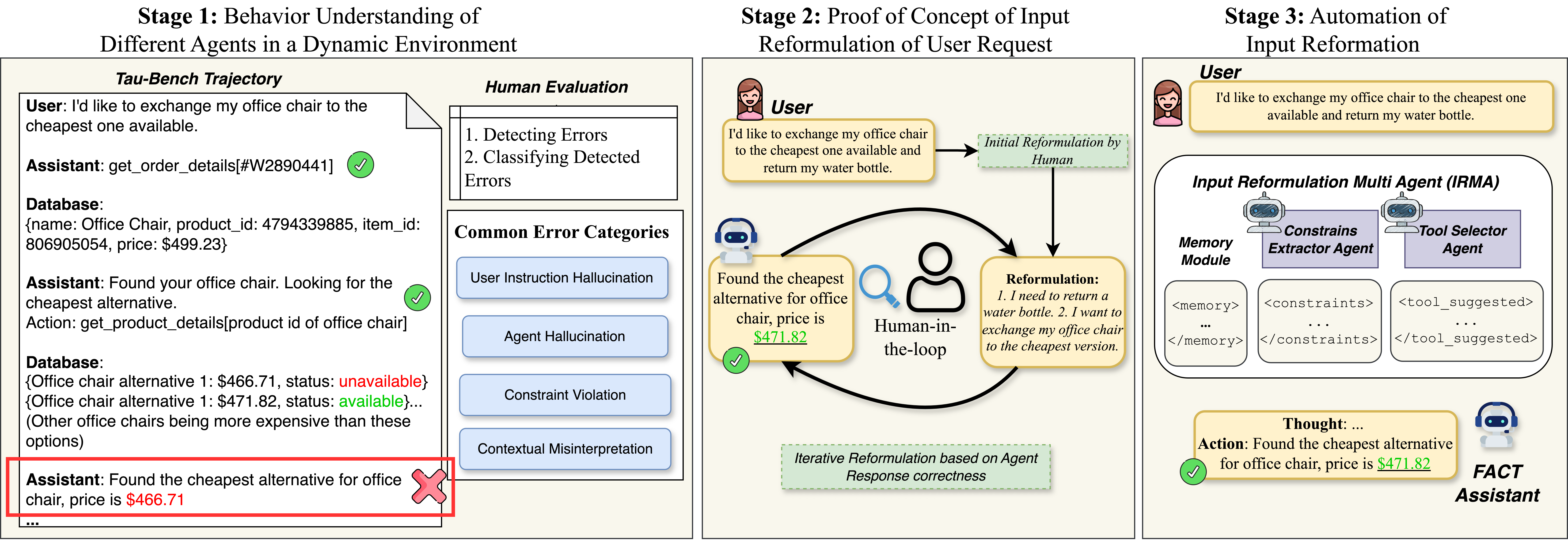}
    \caption{Overview of the tasks conducted for evaluating and improving tool-calling capabilities of language agents in $\tau$-bench \citep{yao2024tau}. \textbf{Stage 1)} involved human evaluators manually evaluating simulated conversation trajectories to find common failure modes of the language agents. \textbf{Stage 2)} employs a human-in-the-loop approach to experiment with various prompt reformulations to improve agent correctness. \textbf{Stage 3)} automates this process through the IRMA framework, which leads to improved agent behavior.}
    \label{fig:teaser}
\end{figure*}

Recent advancements in Large Language Models (LLMs) \citep{annepaka2025large} have created the potential for them to be used as autonomous agents in complex real-world tasks like travel-booking, customer-support, and enterprise operations \citep{chen2024travelagent, wang2024survey, singh-etal-2024-personal, yang2024multi}. However, such complex tasks require the need of reasoning and planning capabilities beyond just language processing: they require the ability on behalf of these agents to be able to invoke suitable tools\footnote{The terms 'tool-use', 'tool-calling' and 'function-calling' are used interchangeably in this paper} which can complete tasks through logic implemented in computer programs leading to deterministic outcomes. Recent research \citep{yao2024tau, lu2024toolsandbox, berkeley-function-calling-leaderboard}, which benchmarks the simulation of such real-world problem-solving settings, shows that LLM-agents significantly falter in correctly solving these tasks and commit errors that range from generative hallucinations to failure to adhere to context and domain-specific policy violations by incorrect reasoning about actions over extended interactions.      

These shortcomings underscore the need for more fine-grained evaluations and methods that can diagnose and address the nuanced failure modes of LLM agents in complex, real-world interactions that employ natural language as a form of communication. Thus, our main focus in this work is to find and mitigate the causes of why language agents fail to solve simulations of real-world conversational requests that require complex reasoning and relevant information processing according to the situation at hand. To this end, we utilize $\tau$-~bench \citep{yao2024tau} as an appropriate test-bed for such investigation as it emulates realistic airline and retail dialogues. We define the reasoning about actions of language agents as the ability to generate context-aware inference and decision-making tokens for selecting the next best action (a tool-call in this context). Additionally, we define and evaluate the planning capabilities of the agents through decision-making for tool-calling over multiple tool-calls in the correct sequential manner to complete a goal. 

Inspired by recent work in context engineering \citep{mei2025survey}, we propose a three-pronged sequential approach. \textit{First}, we develop a comprehensive error classification that categorizes common reasoning and planning mistakes in a multi-turn tool-calling simulation. This taxonomy serves as a diagnostic guideline to systematically identify and understand the causes of failures for LLM agents. \textit{Second}, we manually experiment with input reformulations of the user requests to evaluate whether the correct prompt reformulations can guide the tool-calling agents towards correct decision-making through appropriate tool-calling/response to the user. \textit{Third}, we automate this prompt-reformulation process by building a multi-agent LLM framework (\textsection\ref{subsec:IRMA}), called \textbf{Input-Reformulation Multi-Agent (IRMA)},
which further optimizes the input reformulation with augmentation of follow-up questions (\textsection\ref{subsec:FACT}). Before the tool‑calling agent invokes or responds to any tool output, our automated framework supplies targeted guidance that ensures strict adherence to domain‑specific rules and well-placed follow-up questions to extract accurate information, thereby enhancing its reasoning and planning capabilities in dynamic environments. 

Our results show that the IRMA framework not only outperforms ReAct \cite{yao2023react}, Function Calling, and Self-Reflection \cite{renze2024self} on $\text{pass@1}$, but also achieves\textbf{ 20\%} and \textbf{22.4\%} higher accuracy on Airline tasks compared to Gemini 1.5 Pro-FC and Claude 3.5 Haiku-FC, respectively. IRMA also demonstrates stronger reliability, with higher scores on pass\textasciicircum4 and pass\textasciicircum5 (Figure~\ref{fig:analysis_pass_hat_five}). In addition, IRMA solves tasks in fewer turns than competing methods, highlighting its efficiency (Figure~\ref{fig:turns_analysis}). Lastly, IRMA shows greater robustness, with an increased performance gap on pass\textasciicircum5 after removing tasks affected by ground truth and instruction errors in the airline and retail domains.

The main contributions of our work are:

\begin{enumerate}[noitemsep]
    \item Fine‑grained causal-centric error classification of failure modes occurring in a multi-turn tool-use conversational benchmark.
    
    \item We propose the \textbf{Input-Reformulation Multi-Agent Framework (IRMA)}, a verification-loop-free approach that improves function-calling agents by reformulating prompts with structured and contextually relevant information. IRMA guides the agent to better follow domain policies by enriching its input with key constraints and tool-related context.

    \item We perform an in-depth evaluation of IRMA’s performance across reliability, consistency, and accuracy. Furthermore, our analysis of efficiency reveals that IRMA is able to solve tasks using fewer interaction turns than competing methods.

\end{enumerate}
\section{Related Works}
\label{sec:related_works}

\paragraph{Tool-Integration for LLMs}

The ReAct framework, introduced by \citet{yao2023react}, is one of the first approaches to explore the potential of Large Language Models (LLMs) as tool-using agents by integrating reasoning and acting within LLMs. Toolformer \citep{schick2023toolformer} presents a fine-tuning approach to teach LLMs to invoke tool calls. ToolEVO \citep{chen2024learning} and ToolLLM \citep{qin2023toolllm} employ tree search algorithms for integrating and evaluating tool-learning capabilities in LLMs. 
ToolACE \citep{liu2024toolace}, AutoTools \citep{shi2025toollearningwildempowering}, and APIGen \citep{liu2024apigen} introduce automated frameworks designed to generate accurate, complex, and high-quality tool-learning data, with works like  \citep{prabhakar2025apigen, yin2025magnet} extending this to multi-turn interactive conversational settings. 

\paragraph{Tool-Use Benchmarks}

LLMs have been extensively evaluated on invoking external functions in both single‑turn and interactive multi‑turn conversational test beds. API-Bench \citep{patil2024gorilla} and API-Bank \citep{li2023api} are two prominent benchmarks designed to evaluate the function-calling capabilities of LLMs in single-turn scenarios. NESTful \citep{basu2024nestful} focuses on evaluating LLMs' ability to handle nested sequences of API calls. ToolQA‑D \citep{chen2024learning} gauges robustness in changing API specifications. $\tau$‑bench \citep{yao2024tau} and ToolSandbox 
\citep{lu2024toolsandbox} emulate realistic dialogues requiring policy‑compliant tool use over multi-turn user-agent interactions, where each step modifies an external environment. While these existing multi-turn benchmarks evaluate the overall success of tool-calling agents, they lack fine-grained analysis of reasoning errors while following complex domain rules—a gap our work addresses through the construction of a fine-grained error classification by evaluating $\tau$‑bench.

\paragraph{Improving LLM Tool-Use}
Recent research has explored diverse strategies to enhance the tool-use capabilities of LLMs, focusing on API calling and web-environment interaction—by leveraging techniques such as synthetic data generation, reinforcement learning, and memory augmentation. \citet{liu2024apigen} introduces APIGen, an automated pipeline that generates high-quality, verifiable single-turn function-calling datasets, enabling small models to outperform GPT-4 on the BFCL \citep{patil2025bfcl}. APIGen-MT \citep{prabhakar2025apigen} extends the framework to show improvement in models on multi-turn scenarios through blueprint-driven simulation of human–agent dialogues. ReTool \citep{feng2025retool} integrates dynamic code execution within the reasoning process and training via outcome-driven RL, which significantly improves multi-step reasoning. Nemotron-Tool-N1 \citep{zhang2025nemotron} uses an RL framework to teach precise tool invocation and explicit reasoning, achieving state-of-the-art on API-Bank \citep{li2023api} and BFCL. ARTIST \citep{singh2025agentic} integrates agentic reasoning with RL, enabling LLMs to decide autonomously when and how to call tools. Memento \citep{zhou2025mementofinetuningllmagents} employs a memory-augmented, case-based planner for continual adaptation without retraining, achieving strong generalization on GAIA \citep{mialon2023gaia} and DeepResearcher \citep{zheng2025deepresearcher} benchmarks. While these works mark a shift toward adaptive, planning-driven, and memory-augmented LLM agents by leveraging training methods, our proposed IRMA framework explores tool-use improvement from the perspective of context engineering \citep{mei2025survey} principles.




\section{Problem Statement}
\label{sec:problem_statement}



To evaluate the tool-usage capabilities of current Large Language Models (LLMs), we adopt the benchmark provided by $\tau$-bench \citep{yao2024tau}. This benchmark is specifically designed to assess language agents in realistic, multi-turn interaction settings. $\tau$-bench includes tasks from two domains: (1) Airline, comprising 50 tasks centered around flight reservation scenarios, and (2) Retail, containing 115 tasks focused on shopping and order management. In this setup, both the user and the customer-service assistant are simulated by LLMs, enabling a controlled environment for analyzing interactive behavior. The customer-service agent is the language agent that generates the tokens signifying which tools are to be invoked, while following the specific domain policies (refer Appendix \ref{sec:domain_policies})

Each task is framed as a Partially Observable Markov Decision Process (POMDP) (Details in Appendix \ref{sec:tau-bench-task}), where the assistant agent must generate appropriate function calls based on user inputs. These function calls are executed in an external environment, which then returns outputs that shape the ongoing dialogue. The interaction continues until the user ends the conversation, and the performance of the assistant is evaluated based on final rewards. These rewards reflect how closely the agent's actions align with gold-standard trajectories and how well it fulfills the user’s goals.




A key challenge in $\tau$-bench arises from the dynamic nature of user-agent interactions, where both user inputs and agent responses can vary across runs. This variability requires the agent to consistently execute correct action sequences, regardless of the conversational path. However, current results indicate that even state-of-the-art LLMs struggle to reliably complete these tasks as the number of trials increases. To address this limitation, we conduct a root-cause analysis of common agent errors (\textsection \ref{sec:error-taxonomy}) and introduced IRMA, a multi-agent framework (\textsection \ref{sec:method}) designed to improve agent reliability in this challenging setting.


\section{Error-Classification}
\label{sec:error-taxonomy}
To identify the failure modes of LLMs, human evaluators conducted experiments using GPT-4o \citep{hurst2024gpt} as the base model for both the user and the assistant agent across all tasks in $\tau$-bench \citep{yao2024tau}. Both ReAct and function-calling agent configurations were used to generate up to five trials per task in each domain. Evaluators manually reviewed the resulting multi-turn conversation trajectories from the retail and airline domains. While prior studies \citep{sun2024tools, WinstonJ2025, cemri2025multi} have examined failures related to tool availability, definition errors, or tool set complexity, our analysis focuses specifically on the contextual reasoning limitations of LLMs in generating tool calls within dynamic, multi-turn interactions.

Although $\tau$‑bench provides a general taxonomy of failure types for the retail domain, our classification is more cause-oriented than effect-oriented. By framing errors in terms of their underlying causes, we can more effectively inform the design of targeted interventions, such as retrieval-augmented memory to mitigate context retention issues or follow-up question generation (\textsection \ref{subsec:FACT}) to reduce hallucinations from context drift. The following subsections (\textsection\ref{subsec: user-instruction-hallucination}–\textsection\ref{subsec:context_misinterpret}) provide a detailed breakdown of the identified error types.

\subsection{User Instruction Hallucination}
\label{subsec: user-instruction-hallucination}

User instruction errors occur when the LLM-simulated user deviates from the original task instruction, typically in the later stages of a conversation. These errors highlight the limitations of LLMs in maintaining instruction fidelity over long contexts, especially when multiple follow-up turns introduce competing directives. Another contributing factor is context drift, where the model increasingly relies on recent inputs or high-probability continuations, leading it to overlook or forget the initial user intent. An Example illustrating this error is provided in Figure \ref{fig:task-19-traj-1} in Appendix \ref{sec:failure_examples}.


\subsection{Agent Hallucination}
\label{subsec:agent_hallu}

Agent hallucination errors arise when the assistant agent generates incorrect or incomplete responses that fail to fully satisfy the user’s request. For example, the agent may neglect to process all items specified by the user or incorrectly fulfill a request by selecting the wrong item or applying it to the wrong order. These errors reflect underlying challenges with LLM memory limitations \citep{shan2025cognitive} and the degradation of instruction-following abilities over long contexts \citep{liu2023lost}. As prior context accumulates, excessive or outdated information can distort the model’s understanding, leading to hallucinated outputs and ultimately incorrect decisions \citep{zhang-etal-2024-toolbehonest}.

\subsection{Domain Policy Violation}

Domain policy violations occur when tool-calling agents make decisions that contradict the domain-specific constraints defined for task completion. For instance, in Retail task 19 (Figure \ref{fig:task-19-traj-2}), the agent attempts to exchange the user's office chair and pet bed even when the order is no more in `delivered' status: a prerequisite domain rule required to be satisfied for exchange. This leads to the agent violating the domain rule (see Figure \ref{fig:retail-domain-part-2}): '\textit{An order can only be exchanged if its status is 'delivered'...}' Such violations may also arise when the user issues an invalid request, and the agent proceeds to fulfill it without adhering to the applicable domain rules. This error is caused due to similar reasons as mentioned in \textsection \ref{subsec: user-instruction-hallucination} and \textsection \ref{subsec:agent_hallu}.

\subsection{Contextual Misinterpretation}
\label{subsec:context_misinterpret}


Contextual misinterpretation errors occur when the tool-calling agent misunderstands the intent or nuance of the user's request and generates function calls using inappropriate tools for the given context. For example, if a user asks to \textit{return} an item and receive a different one in exchange, a human familiar with the domain policies would recognize this as an \textit{exchange} request. However, the LLM-based agent may misinterpret it as a simple \textit{return}, failing to grasp the full context and thereby invoking the wrong tool.

\section{Method}
\label{sec:method}
As outlined in the previous sections, complex dynamic environments such as $\tau$‑bench present reliability challenges. Specifically, the user simulator may hallucinate during interactions, generating questions that do not adhere to the provided instructions. In this study, we aim to improve the assistant agent's tool-calling performance in $\tau$‑bench by enabling more accurate decision-making. Unlike prior approaches that monitor and correct agent actions through verification or reflection, our method focuses on enhancing the quality of the agent’s input before any action is taken. To achieve this, we first introduce a novel prompting strategy: Follow-up Question Acting (FACT), designed to support decision-making in dynamic settings. We then present the Input Reformulation Multi-Agent (IRMA) framework that reformulates the agent’s input to guide more effective and context-aware decisions.

\begin{figure}
    \centering
    \includegraphics[width=1\linewidth]{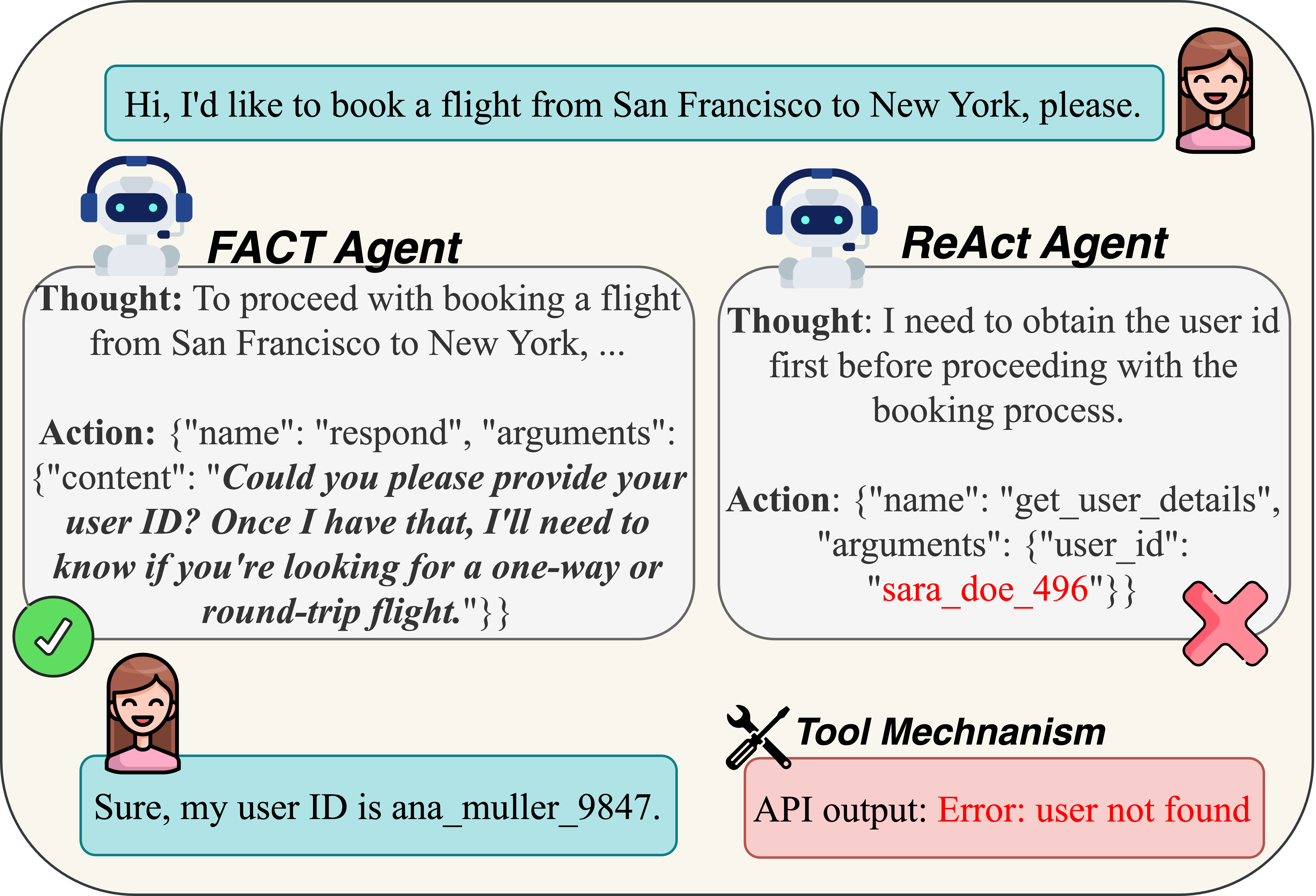}
    \caption{FACT agent demonstrates superior user guidance, avoiding tool-call errors encountered by the ReAct Agent.}
    \label{fig:fact_agent}
\end{figure}

\subsection{FACT: Follow-up question ACTing}
\label{subsec:FACT}
Although reasoning-based prompting techniques like ReAct outperform non-reasoning methods such as Act, they remain inefficient in dynamic environments. As shown in Figure \ref{fig:fact_agent}, ReAct often calls a tool prematurely, triggers an error, and only then asks clarifying questions, leading to longer conversations and increased interaction issues. To overcome this, we introduce \textbf{Follow-up Question ACTing (FACT)}, a prompting method that first gathers information through targeted questions before calling a tool. Our results in Figure \ref{fig:turns_analysis} show that FACT is more effective than ReAct and performs comparably to Function Calling. We refer readers to Appendix \textsection \ref{sec:appendix_fact}.

Another advantage of FACT is its ability to involve the user in the loop. When the user simulator hallucinates or provides misleading input, FACT detects the issue and hands off the conversation to a human, ensuring more robust handling of unreliable inputs. In summary, FACT is more efficient, reliable, and consistent than other methods in dynamic environments. However, in long conversations, it may forget domain rules and tools due to system prompt limitations, leading to domain violations. To address this, we propose the Input-Reformulation Multi-Agent Framework (IRMA), which restructures the user prompt to retain key information like domain rules and a relevant tool list within the assistant's input.


\subsection{IRMA: Input-Reformulation Multi-Agent Framework}
\label{subsec:IRMA}
Our analysis reveals three key failure cases for assistant agents. First, in long conversations, the agent may forget the user’s initial request and respond only partially. Second, it may violate domain rules by forgetting constraints from lengthy policy lists. Third, tool selection becomes harder over time, especially when tools have similar names (e.g., "search\_direct\_flight" vs. "search\_onestep\_flight"), leading to incorrect choices.



We hypothesize that combining user queries with crucial context, such as domain rules and relevant tools, can improve the assistant agent’s decision-making. To test this, we conducted a \textit{human-in-the-loop} experiment with prompt engineers who reformulated queries using additional policy and tool information. In most cases, the agent successfully completed the tasks, motivating us to automate this input reformulation process.


Based on this insight, we propose the \textbf{Input Reformulation Multi-Agent Framework (IRMA)}. In contrast to prior methods that focus on post-hoc correction of the agent’s behavior, such as Self-Reflection, PlanGen \cite{parmar2025plangen}, or other verification-based approaches, IRMA centers on enhancing the quality of the input provided to the assistant agent. This approach enhances decision-making at the input stage—before any action is taken—ensuring more accurate and context-aware responses. The framework comprises three core modules: memorization, constraints, and tool suggestion.

\paragraph{Memorization} This module is independent of the language model and is responsible for storing the user queries throughout the interaction trajectory. It helps the agent retain awareness of the initial request and make decisions accordingly. The conversation history is maintained within \texttt{<memory>} tags.

\paragraph{Constraints} One of the main reasons the agent makes incorrect decisions is domain policy violation. A key insight from the human-in-the-loop experiment was the positive impact of providing a concise list of domain constraints to guide the assistant agent’s decisions. To address this challenge, we define a dedicated agent that generates a checklist of relevant domain constraints based on the user query. If the user query is a response to a follow-up question from the assistant, the agent is prompted to return “None”. The generated constraint list is stored within \texttt{<constraints>} tags to ensure the assistant agent receives a structured and interpretable input prompt.

\paragraph{Tool Suggestion} Although the number of available tools is limited, the assistant agent sometimes struggles to select the most relevant tool for a given user query. In some cases, after encountering an error or receiving an empty output, the agent may lose track of other parts of the user’s request. To mitigate this, we introduce a Tool Selector agent that generates a short list of tools most relevant to the user query, along with a one-line explanation for each suggestion. This list is stored within \texttt{<tool\_suggested>} tags to help the assistant agent focus on selecting the most appropriate tool.

In summary, the IRMA framework aims to replicate the input reframing performed by researchers during the human-in-the-loop experiment. Unlike other techniques such as verification, self-reflection, or agentic verification methods, IRMA functions in a loop-free manner and focuses on strengthening the input by reformulating the user query. This approach not only improves accuracy but also offers better cost-effectiveness compared to alternative methods. In the next section, we provide a comparative analysis of IRMA against existing techniques.

\begin{table}[t]
\centering
\small
\resizebox{0.98\linewidth}{!}{
\begin{tabular}{l|c|c|c|c}
\toprule
\textbf{Model} & \textbf{Method} & \textbf{$\boldsymbol{\tau}$-Retail}
 & \textbf{$\boldsymbol{\tau}$-Airline} & \textbf{Overall}
 \\
\midrule
\multicolumn{5}{c}{\textit{Open-Source Models}} \\ \midrule
Qwen 2.5 32B & ReAct & 24.4 & 25.0 & 24.7 \\
Llama 3.1 70B & ReAct & 50.4 & 26.0 & 38.2 \\
DeepSeek v3$^1$ & ReAct & 58.3 & 22.8 & 40.6 \\
Phi-4 14B & ReAct & 32.2 & 28.0 & 30.1 \\
\midrule
\multicolumn{5}{c}{\textit{Close-Source Models}} \\ \midrule
Gemini 1.5 pro$^1$& FC &	54.9 &	25.2 & 40.1 \\
Claude 3.5 Haiku$^2$& FC & 51.0 & 22.8 & 36.9 \\
Claude 3.5 Sonnet$^2$& FC & \textbf{62.6} & 36.0 & \underline{49.3} \\
gpt-4o & FC & \underline{60.5} & 42.4 & 51.4 \\
gpt-4o & ReAct & 51.8 & 39.6 & 45.7 \\
gpt-4o & SR & 51.1 & \underline{44.8} & 47.9 \\
\midrule
gpt-4o \textbf{(ours)} & IRMA & 58.3 & \textbf{47.2} & \textbf{52.75} \\

\bottomrule
\end{tabular}}
\caption{Performance of various open and closed-source models in Pass\textasciicircum 1 for retail and airline domains in $\tau$-bench across 5 runs. 'SR' stands for the Self-Reflection agent. \textsuperscript{1} indicates results from \citep{scaledcognition2025apt1}; \textsuperscript{2} indicates results from \citep{anthropic2024claude35}}
\label{tab:pass_one_results_tables}
\end{table}

\begin{figure*}[t]
    \centering
    \includegraphics[width=1\linewidth]{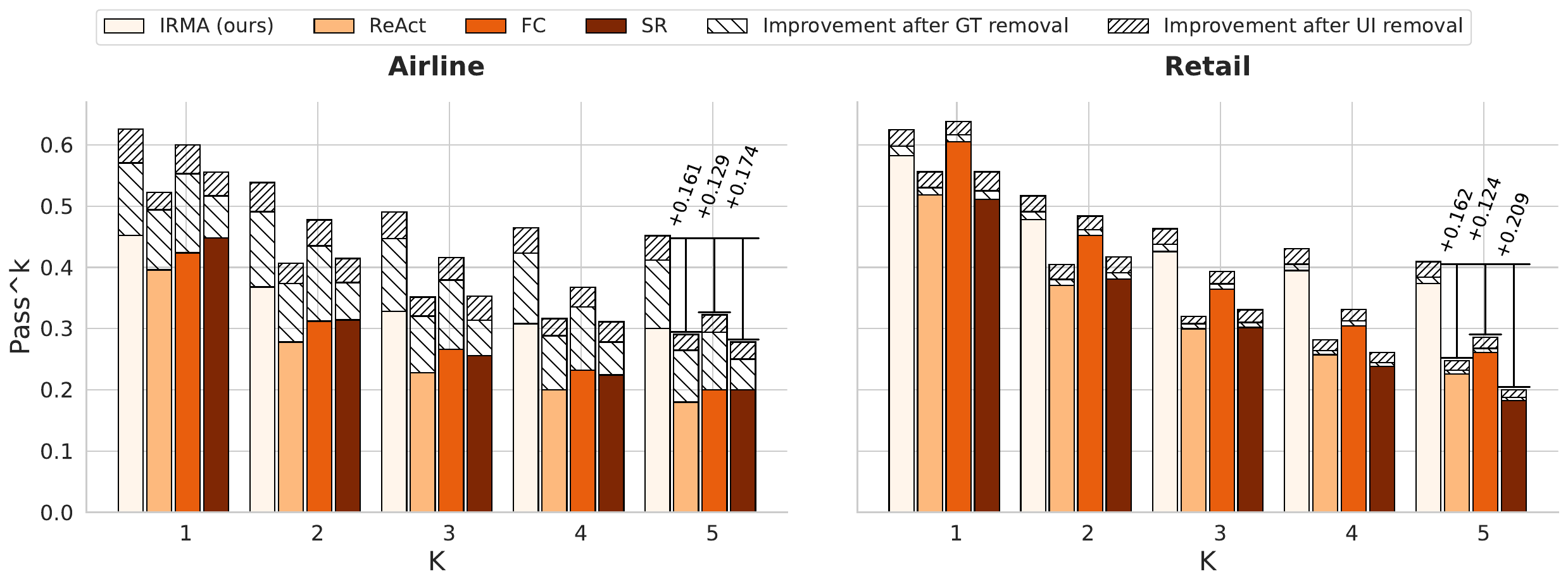}
    \caption{Comparison of IRMA and other techniques across five runs with varying values of K. The figure shows a significant performance difference between IRMA and other methods on pass\textasciicircum5. Note that all methods use GPT-4o as the base model. See Appendix \ref{sec:app_pass_hat_k_results} for more details.}
    \label{fig:analysis_pass_hat_five}
\end{figure*}

\section{Experiments}
\label{sec:experiments}
\subsection{Experimental Setup}
We present the baseline models and comparison methods, followed by an analysis of the IRMA framework using various evaluation metrics and ablation studies (refer Appendix \ref{sec:ablation_on_IRMA}) to assess the impact of its individual components on $\tau$-bench performance.

\paragraph{Models and Methods} We evaluated IRMA against a range of open-source and closed-source language models. The open-source models include Qwen2.5-32B \citep{qwen2025qwen25technicalreport}, LLaMA-3.1-70B \citep{grattafiori2024llama}, DeepSeek-v3 \citep{liu2024deepseek}, and Phi-4-14B \citep{abdin2024phi}, while the closed-source models comprise Claude 3.5 \citep{anthropic2024claude3.5sonnet}, Gemini 1.5 \citep{team2024gemini}, and GPT-4o \citep{hurst2024gpt}. In addition, we compared IRMA with three widely adopted prompting strategies: (1) ReAct, a reasoning-based prompting technique; (2) Function Calling, designed specifically to enhance a model’s tool-calling capability; and (3) Self-Reflection, a method aimed at improving tool-use performance by addressing errors in the agent's actions.

\paragraph{Evaluation} 
To evaluate performance, we use the pass\textasciicircum k metrics, which measure the reliability and consistency of models across different prompting strategies. The pass\textasciicircum k metric (pronounced "pass hat k") is defined as the probability that all of the k independently sampled outputs successfully complete the task, averaged across all tasks. Specifically, if a task is run for $n$ independent trials and $c$ of those are successful (i.e., have a correct result with reward $r=1$), an unbiased estimate of pass \textasciicircum k can be computed using the following formula:

\begin{equation*}
    \text{pass\textasciicircum k} = \mathbb{E}_{\text{task}}\left[ \binom{c}{k} \middle/ \binom{n}{k} \right]
\end{equation*}

This metric provides insight into how likely a model is to succeed given multiple attempts, capturing both reliability and diversity in its outputs.


\subsection{Experimental Results}
\label{sec:experimental-results}
As outlined in the $\tau$-bench, in real-world scenarios—reliability and consistency are often more critical than the average success rate (measured by $\text{pass@1}$). We argue that an ideal agentic method should exhibit three key properties: (1) Accuracy, (2) Reliability, and (3) Consistency. Accordingly, we begin by comparing results using $\text{pass@1}$ to assess accuracy, and then evaluate the performance of state-of-the-art methods using pass\textasciicircum k to measure reliability and consistency.



\paragraph{IRMA outperforms other state-of-the-art methods in tool calling.}
We conducted evaluations of multiple methods—Function Calling (FC), ReAct, and Self-Reflection—each executed over five trials. These experiments were performed using the GPT-4o model. The results, presented in Table \ref{tab:pass_one_results_tables}, show that the IRMA framework outperforms ReAct, Self-Reflection, and FC by 6.1\%, 3.9\%, and 0.4\%, respectively, in overall $\text{pass@1}$ score. Additionally, in the airline tasks, which represent the most challenging scenarios within the dynamic environment, IRMA on GPT-4o achieves improvements of 20\%, 22.4\%, and 9.2\% compared to Gemini 1.5 Pro-FC, Claude 3.5 Haiku-FC, and Claude 3.5 Sonnet-FC, respectively. These findings highlight IRMA’s strong accuracy in real-world tasks and demonstrate its effectiveness over existing methods.

\begin{figure}[h!]
    \centering
    \includegraphics[width=1\linewidth]{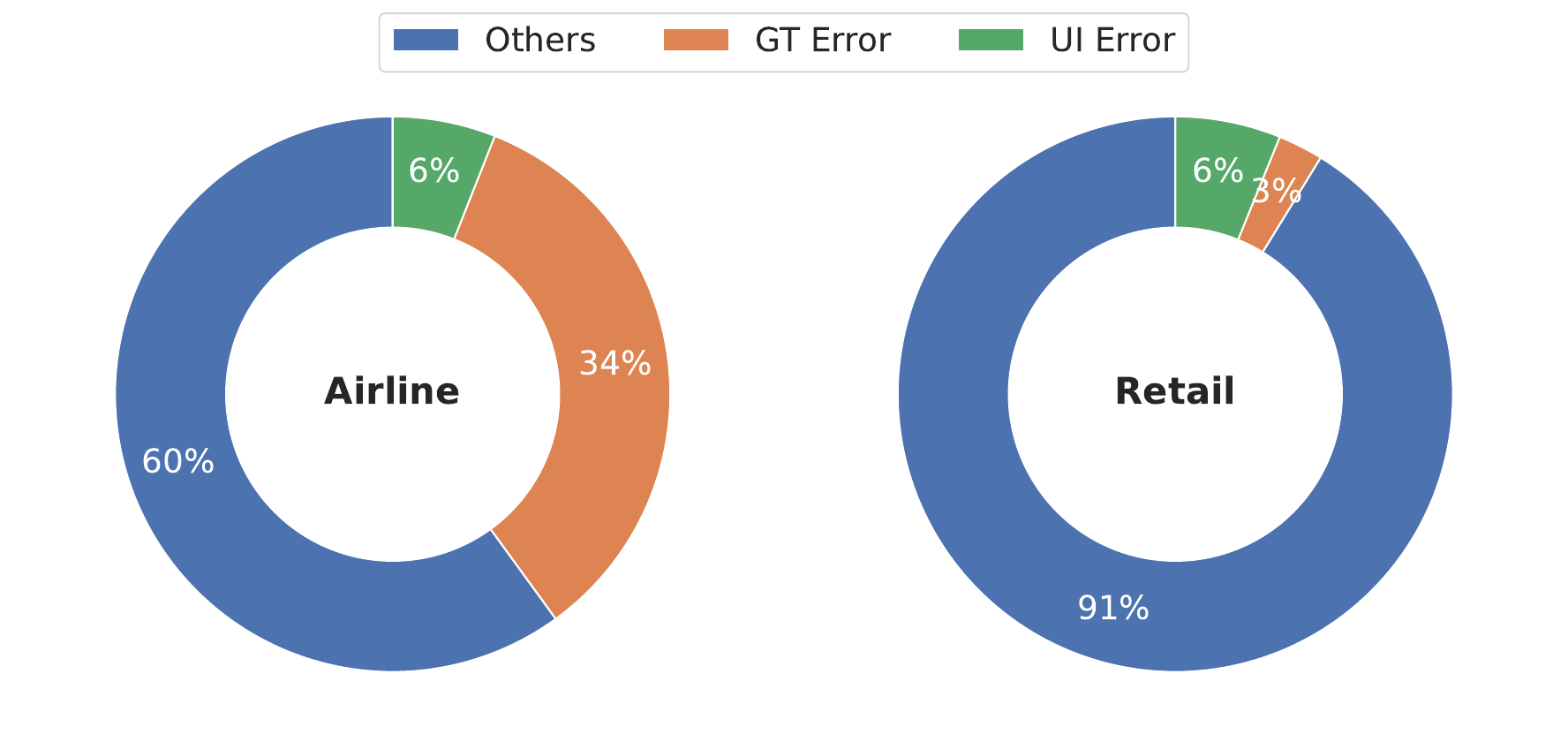}
    \caption{Error statistics across Airline and Retail tasks. \textbf{GT:} Ground Truth errors; \textbf{UI:} User Instruction errors.}
    \label{fig:error_analysis_tasks}
\end{figure}

\begin{figure}
    \centering
    \includegraphics[width=1\linewidth]{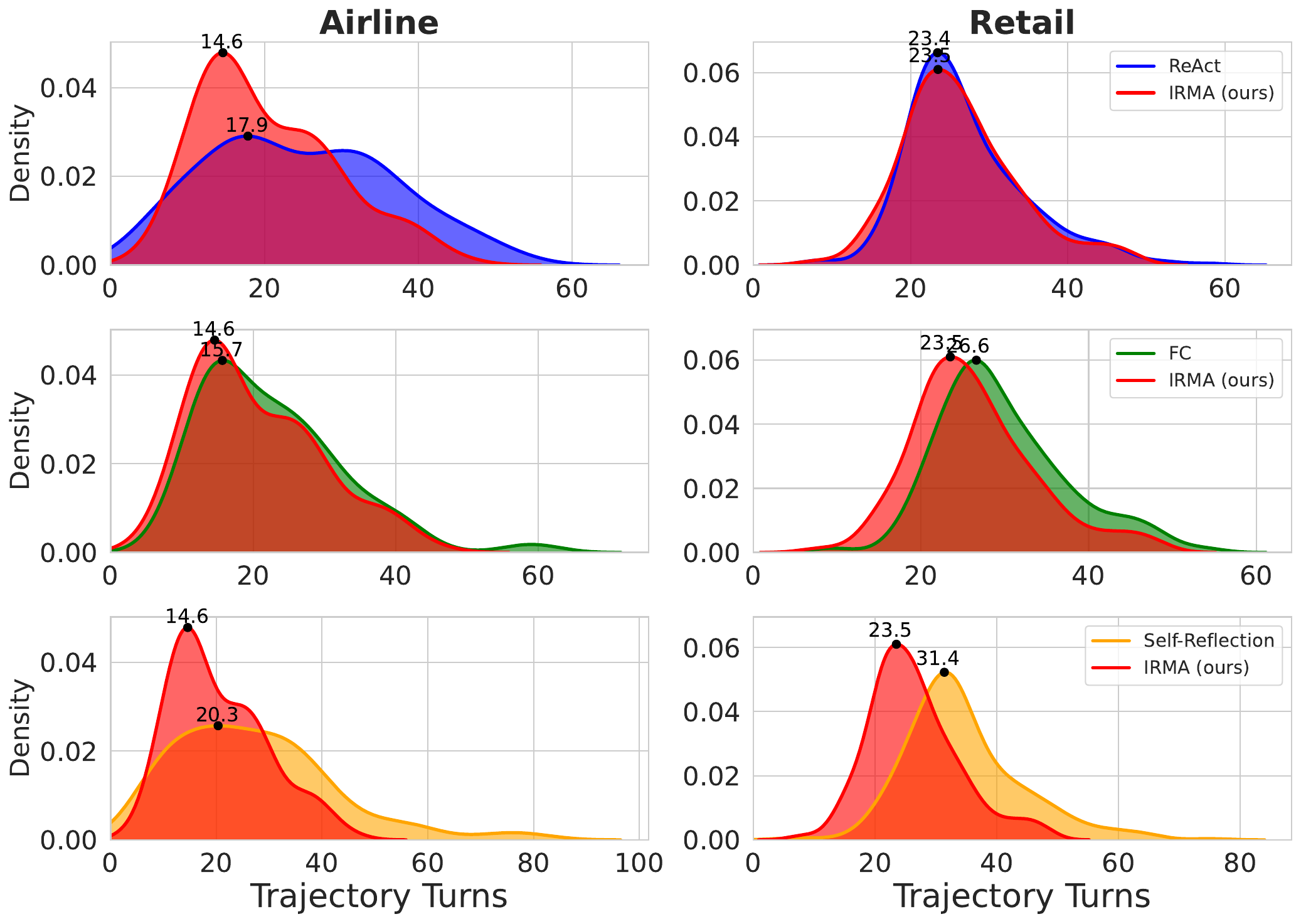}
    \caption{Comparison of IRMA and other methods based on the number of turns in successful tasks in the Airline and Retail domains.}
    \label{fig:turns_analysis}
\end{figure}

\paragraph{IRMA is more reliable and consistent than other methods in dynamic settings.}
The results in Table \ref{tab:pass_one_results_tables} show that the performance of IRMA on retail pass\textasciicircum 1 is slightly lower than that of GPT-4o-FC. For this reason, we further explored the performance of other methods using pass\textasciicircum k to evaluate their reliability and consistency. The results in Figure \ref{fig:turns_analysis} show that IRMA, compared with ReAct and FC on GPT-4o, is much more reliable and consistent, outperforming ReAct and FC by 16.1\% and 12.6\%, respectively, in overall scores on pass\textasciicircum5.

\paragraph{IRMA is more robust on tasks with GT and UI errors.}
As explained in the previous sections, $\tau$‑bench suffers from two major issues: (1) Ground Truth (GT) errors and (2) User Instruction (UI) errors. Figure \ref{fig:error_analysis_tasks} shows the distribution of these errors across the airline and retail tasks. We progressively removed tasks affected by these problems, and the results revealed that the performance of all three methods improved, with IRMA showing slightly greater gains compared to the others. We hypothesize that IRMA is more robust to hallucination-related issues. Specifically, in tasks with GT errors, IRMA tends to avoid incorrect tool calls or invalid actions and instead produces safe and accurate responses.
\\
\\
A key observation is the change in performance difference between IRMA and FC on pass\textasciicircum 5. Before removing tasks with GT and UI errors, IRMA outperformed FC by 10\%. However, after removing these problematic tasks, the performance gap widened to 16.1\% on average. Similar patterns were observed for other methods as well, reinforcing the claim that IRMA is more robust and less sensitive to noisy supervision and ambiguous instructions compared to existing techniques.

\paragraph{IRMA solves tasks more efficiently and effectively, using fewer turns than others.}
One of the primary reasons assistant agents make incorrect decisions in the final turns is the length of the conversation, which often causes them to forget important rules and instructions. In an ideal scenario, an assistant should resolve the user's query with the fewest but most effective actions. To investigate this aspect, we analyzed the distribution of turns in successful task completions by IRMA, ReAct, FC, and Self-Reflection, as shown in Figure \ref{fig:turns_analysis}. The results show that, in retail tasks, IRMA completes tasks with 7.9 points fewer turns than Self-Reflection and 3.1 points fewer than FC. In airline tasks, IRMA requires 8.3 fewer turns than Self-Reflection, 1.1 fewer than FC, and 3.3 fewer than ReAct. These results demonstrate IRMA’s superior efficiency compared to other state-of-the-art methods.


\paragraph{Input Reformulation framework vs Self-Reflection} The central concept of IRMA is to reformulate the agent’s input under the assumption that supplying sufficient and well-structured information enables the agent to act more reliably and consistently in real-world scenarios. To evaluate this, we implemented the Self-Reflection method (Appendix \ref{sec:self-reflect-agent}), which analyzes the agent’s previous actions and extracts relevant information from domain rules to guide future decisions (see section \ref{sec:appendix_fact} for implementation details). As shown in Figure \ref{fig:analysis_pass_hat_five}, IRMA outperforms Self-Reflection in both airline and retail tasks, achieving a 3.9\% higher overall score in $\text{pass@1}$. More notably, IRMA exceeds Self-Reflection by 19.1\% in pass\textasciicircum 5, highlighting its superior reliability in a real-world environment.
\\
\\
In summary, while ReAct and Self-Reflection perform well in certain settings, they fall short in complex, dynamic environments like $\tau$‑bench. Role-play methods, including verification techniques, are also inefficient, as real-world scenarios require assistant agents to act based on limited information, with each action affecting the environment. Although Function Calling was designed for tool use, our results show it lacks reliability in decision-making and offers limited controllability, even in GPT-4o with tailored system prompts. Combining FACT with GPT-4o-FC led to a 12\% performance drop, highlighting the need for more robust approaches. In contrast, IRMA consistently delivers higher accuracy, reliability, and consistency in dynamic environments like $\tau$‑bench.

\section{Conclusion}
\label{sec:conclusion}
In this work, we investigate the limitations of state-of-the-art LLM-based tool-calling agents in complex, multi-turn environments, focusing on the retail and airline domains of $\tau$-bench. Through a detailed analysis of conversation trajectories, we identify four major failure modes: \textit{user instruction hallucination, agent hallucination}, \textit{domain-policy violations}, and \textit{contextual misinterpretation}, all of which stem from limitations in memory retention, contextual reasoning, and adherence to domain constraints across extended interactions. To address these challenges, we propose the Input Reformulation Multi-Agent (IRMA) framework, designed to enhance the structure of the assistant agent’s input. Our results show that IRMA not only outperforms other methods in pass\textasciicircum 1 but also demonstrates significantly higher reliability, achieving an overall score of 43\%  pass\textasciicircum 5 in $\tau$-bench. Moreover, by leveraging the FACT agent, IRMA exhibits greater efficiency in task completion. In conclusion, IRMA shows robust and consistent behavior in the unreliable and dynamic environment of $\tau$-bench, highlighting its effectiveness in real-world tool-use scenarios.

\section*{Limitations}
\label{sec:limitations}
Although the Input Reformulation Multi-Agent (IRMA) framework demonstrated superior performance on $\tau$-bench, several limitations remain. As shown in Figure \ref{fig:analysis_pass_hat_five}, while IRMA exhibits greater reliability compared to other methods, its performance on pass\textasciicircum5 still hovers around 43\%. This indicates that there is still considerable room for improving the reliability of tool-using agents in real-world scenarios. Another limitation of this work is that our experiments and analysis are restricted to the $\tau$-bench benchmark. It would be valuable to evaluate IRMA across a broader range of real-world environments to assess its generalizability.
\\
\\
Moreover, our observations suggest that beyond the error taxonomy we proposed, $\tau$-bench itself suffers from issues related to unfair reward modeling. Building a truly dynamic and reliable evaluation environment—especially one that can control for the correctness of user instructions—would have a significant impact on the field. Such an environment would enable more rigorous development and evaluation of agentic frameworks and encourage further research into robust, real-world agent behavior. Ultimately, we believe this work contributes meaningfully to the research community and provides a strong foundation for developing more reliable and consistent agentic methods for dynamic environments.

\section*{Ethics Statement}
\label{sec:ethics statement}
We have utilized AI assistants, specifically Grammarly
and ChatGPT, to correct grammatical errors and
rephrase sentences.

\section*{Acknowledgement}

We thank the anonymous reviewers for their constructive suggestions. We extend our gratitude to the Research Computing (RC), and Enterprise Technology at ASU for providing computing resources, and access to the GPT API version for experiments. This work was in part supported by a gift award from Cisco Research.
\bibliography{references}

\begin{thebibliography}{45}
\providecommand{\natexlab}[1]{#1}

\bibitem[{Abdin et~al.(2024)Abdin, Aneja, Behl, Bubeck, Eldan, Gunasekar, Harrison, Hewett, Javaheripi, Kauffmann et~al.}]{abdin2024phi}
Marah Abdin, Jyoti Aneja, Harkirat Behl, S{\'e}bastien Bubeck, Ronen Eldan, Suriya Gunasekar, Michael Harrison, Russell~J Hewett, Mojan Javaheripi, Piero Kauffmann, et~al. 2024.
\newblock Phi-4 technical report.
\newblock \emph{arXiv preprint arXiv:2412.08905}.

\bibitem[{Annepaka and Pakray(2025)}]{annepaka2025large}
Yashwanth Annepaka and Prasenjit Pakray. 2025.
\newblock \href {https://doi.org/10.1007/s10115-024-02310-4} {Large language models: a survey of their development, capabilities, and applications}.
\newblock \emph{Knowledge and Information Systems}, 67:2967--3022.

\bibitem[{{Anthropic}(2024{\natexlab{a}})}]{anthropic2024claude35}
{Anthropic}. 2024{\natexlab{a}}.
\newblock Claude 3.5 models and computer use.
\newblock \url{https://www.anthropic.com/news/3-5-models-and-computer-use}.
\newblock Accessed: 2025-05-20.

\bibitem[{{Anthropic}(2024{\natexlab{b}})}]{anthropic2024claude3.5sonnet}
{Anthropic}. 2024{\natexlab{b}}.
\newblock Claude 3.5 sonnet.
\newblock \url{https://www.anthropic.com/news/claude-3-5-sonnet}.
\newblock 4 min read.

\bibitem[{Basu et~al.(2024)Basu, Abdelaziz, Kate, Agarwal, Crouse, Rizk, Bradford, Munawar, Kumaravel, Goyal et~al.}]{basu2024nestful}
Kinjal Basu, Ibrahim Abdelaziz, Kiran Kate, Mayank Agarwal, Maxwell Crouse, Yara Rizk, Kelsey Bradford, Asim Munawar, Sadhana Kumaravel, Saurabh Goyal, et~al. 2024.
\newblock Nestful: A benchmark for evaluating llms on nested sequences of api calls.
\newblock \emph{arXiv preprint arXiv:2409.03797}.

\bibitem[{Cemri et~al.(2025)Cemri, Pan, Yang, Agrawal, Chopra, Tiwari, Keutzer, Parameswaran, Klein, Ramchandran et~al.}]{cemri2025multi}
Mert Cemri, Melissa~Z Pan, Shuyi Yang, Lakshya~A Agrawal, Bhavya Chopra, Rishabh Tiwari, Kurt Keutzer, Aditya Parameswaran, Dan Klein, Kannan Ramchandran, et~al. 2025.
\newblock Why do multi-agent llm systems fail?
\newblock \emph{arXiv preprint arXiv:2503.13657}.

\bibitem[{Chen et~al.(2024{\natexlab{a}})Chen, Ge, Fu, Xiao, and Chen}]{chen2024travelagent}
Aili Chen, Xuyang Ge, Ziquan Fu, Yanghua Xiao, and Jiangjie Chen. 2024{\natexlab{a}}.
\newblock Travelagent: An ai assistant for personalized travel planning.
\newblock \emph{arXiv preprint arXiv:2409.08069}.

\bibitem[{Chen et~al.(2024{\natexlab{b}})Chen, Zhang, Cong, Guo, Wu, Lin, Feng, and Wang}]{chen2024learning}
Guoxin Chen, Zhong Zhang, Xin Cong, Fangda Guo, Yesai Wu, Yankai Lin, Wenzheng Feng, and Yasheng Wang. 2024{\natexlab{b}}.
\newblock Learning evolving tools for large language models.
\newblock \emph{arXiv preprint arXiv:2410.06617}.

\bibitem[{Cognition(2025)}]{scaledcognition2025apt1}
Scaled Cognition. 2025.
\newblock Apt-1: Adaptive prompt tuning for llms.
\newblock \url{https://www.scaledcognition.com/blog/apt-1}.
\newblock Accessed: 2025-05-19.

\bibitem[{Feng et~al.(2025)Feng, Huang, Qu, Zhang, Qin, Zhong, Jiang, Chi, and Zhong}]{feng2025retool}
Jiazhan Feng, Shijue Huang, Xingwei Qu, Ge~Zhang, Yujia Qin, Baoquan Zhong, Chengquan Jiang, Jinxin Chi, and Wanjun Zhong. 2025.
\newblock Retool: Reinforcement learning for strategic tool use in llms.
\newblock \emph{arXiv preprint arXiv:2504.11536}.

\bibitem[{Grattafiori et~al.(2024)Grattafiori, Dubey, Jauhri, Pandey, Kadian, Al-Dahle, Letman, Mathur, Schelten, Vaughan et~al.}]{grattafiori2024llama}
Aaron Grattafiori, Abhimanyu Dubey, Abhinav Jauhri, Abhinav Pandey, Abhishek Kadian, Ahmad Al-Dahle, Aiesha Letman, Akhil Mathur, Alan Schelten, Alex Vaughan, et~al. 2024.
\newblock The llama 3 herd of models.
\newblock \emph{arXiv preprint arXiv:2407.21783}.

\bibitem[{Hurst et~al.(2024)Hurst, Lerer, Goucher, Perelman, Ramesh, Clark, Ostrow, Welihinda, Hayes, Radford et~al.}]{hurst2024gpt}
Aaron Hurst, Adam Lerer, Adam~P Goucher, Adam Perelman, Aditya Ramesh, Aidan Clark, AJ~Ostrow, Akila Welihinda, Alan Hayes, Alec Radford, et~al. 2024.
\newblock Gpt-4o system card.
\newblock \emph{arXiv preprint arXiv:2410.21276}.

\bibitem[{Li et~al.(2023)Li, Zhao, Yu, Song, Li, Yu, Li, Huang, and Li}]{li2023api}
Minghao Li, Yingxiu Zhao, Bowen Yu, Feifan Song, Hangyu Li, Haiyang Yu, Zhoujun Li, Fei Huang, and Yongbin Li. 2023.
\newblock Api-bank: A comprehensive benchmark for tool-augmented llms.
\newblock \emph{arXiv preprint arXiv:2304.08244}.

\bibitem[{Liu et~al.(2024{\natexlab{a}})Liu, Feng, Xue, Wang, Wu, Lu, Zhao, Deng, Zhang, Ruan et~al.}]{liu2024deepseek}
Aixin Liu, Bei Feng, Bing Xue, Bingxuan Wang, Bochao Wu, Chengda Lu, Chenggang Zhao, Chengqi Deng, Chenyu Zhang, Chong Ruan, et~al. 2024{\natexlab{a}}.
\newblock Deepseek-v3 technical report.
\newblock \emph{arXiv preprint arXiv:2412.19437}.

\bibitem[{Liu et~al.(2023)Liu, Lin, Hewitt, Paranjape, Bevilacqua, Petroni, and Liang}]{liu2023lost}
Nelson~F Liu, Kevin Lin, John Hewitt, Ashwin Paranjape, Michele Bevilacqua, Fabio Petroni, and Percy Liang. 2023.
\newblock Lost in the middle: How language models use long contexts.
\newblock \emph{arXiv preprint arXiv:2307.03172}.

\bibitem[{Liu et~al.(2024{\natexlab{b}})Liu, Huang, Zeng, Hao, Yu, Li, Wang, Gan, Liu, Yu et~al.}]{liu2024toolace}
Weiwen Liu, Xu~Huang, Xingshan Zeng, Xinlong Hao, Shuai Yu, Dexun Li, Shuai Wang, Weinan Gan, Zhengying Liu, Yuanqing Yu, et~al. 2024{\natexlab{b}}.
\newblock Toolace: Winning the points of llm function calling.
\newblock \emph{arXiv preprint arXiv:2409.00920}.

\bibitem[{Liu et~al.(2024{\natexlab{c}})Liu, Hoang, Zhang, Zhu, Lan, Tan, Yao, Liu, Feng, RN et~al.}]{liu2024apigen}
Zuxin Liu, Thai Hoang, Jianguo Zhang, Ming Zhu, Tian Lan, Juntao Tan, Weiran Yao, Zhiwei Liu, Yihao Feng, Rithesh RN, et~al. 2024{\natexlab{c}}.
\newblock Apigen: Automated pipeline for generating verifiable and diverse function-calling datasets.
\newblock \emph{Advances in Neural Information Processing Systems}, 37:54463--54482.

\bibitem[{Lu et~al.(2024)Lu, Holleis, Zhang, Aumayer, Nan, Bai, Ma, Ma, Li, Yin et~al.}]{lu2024toolsandbox}
Jiarui Lu, Thomas Holleis, Yizhe Zhang, Bernhard Aumayer, Feng Nan, Felix Bai, Shuang Ma, Shen Ma, Mengyu Li, Guoli Yin, et~al. 2024.
\newblock Toolsandbox: A stateful, conversational, interactive evaluation benchmark for llm tool use capabilities.
\newblock \emph{arXiv preprint arXiv:2408.04682}.

\bibitem[{Mei et~al.(2025)Mei, Yao, Ge, Wang, Bi, Cai, Liu, Li, Li, Zhang et~al.}]{mei2025survey}
Lingrui Mei, Jiayu Yao, Yuyao Ge, Yiwei Wang, Baolong Bi, Yujun Cai, Jiazhi Liu, Mingyu Li, Zhong-Zhi Li, Duzhen Zhang, et~al. 2025.
\newblock A survey of context engineering for large language models.
\newblock \emph{arXiv preprint arXiv:2507.13334}.

\bibitem[{Mialon et~al.(2023)Mialon, Fourrier, Wolf, LeCun, and Scialom}]{mialon2023gaia}
Gr{\'e}goire Mialon, Cl{\'e}mentine Fourrier, Thomas Wolf, Yann LeCun, and Thomas Scialom. 2023.
\newblock Gaia: a benchmark for general ai assistants.
\newblock In \emph{The Twelfth International Conference on Learning Representations}.

\bibitem[{Parmar et~al.(2025)Parmar, Liu, Goyal, Chen, Le, Mishra, Mobahi, Gu, Wang, Nakhost et~al.}]{parmar2025plangen}
Mihir Parmar, Xin Liu, Palash Goyal, Yanfei Chen, Long Le, Swaroop Mishra, Hossein Mobahi, Jindong Gu, Zifeng Wang, Hootan Nakhost, et~al. 2025.
\newblock Plangen: A multi-agent framework for generating planning and reasoning trajectories for complex problem solving.
\newblock \emph{arXiv preprint arXiv:2502.16111}.

\bibitem[{Patil et~al.(2025)Patil, Mao, Cheng-Jie~Ji, Yan, Suresh, Stoica, and E.~Gonzalez}]{patil2025bfcl}
Shishir~G. Patil, Huanzhi Mao, Charlie Cheng-Jie~Ji, Fanjia Yan, Vishnu Suresh, Ion Stoica, and Joseph E.~Gonzalez. 2025.
\newblock The berkeley function calling leaderboard (bfcl): From tool use to agentic evaluation of large language models.
\newblock In \emph{Forty-second International Conference on Machine Learning}.

\bibitem[{Patil et~al.(2024)Patil, Zhang, Wang, and Gonzalez}]{patil2024gorilla}
Shishir~G Patil, Tianjun Zhang, Xin Wang, and Joseph~E Gonzalez. 2024.
\newblock Gorilla: Large language model connected with massive apis.
\newblock \emph{Advances in Neural Information Processing Systems}, 37:126544--126565.

\bibitem[{Prabhakar et~al.(2025)Prabhakar, Liu, Yao, Zhang, Zhu, Wang, Liu, Awalgaonkar, Chen, Hoang et~al.}]{prabhakar2025apigen}
Akshara Prabhakar, Zuxin Liu, Weiran Yao, Jianguo Zhang, Ming Zhu, Shiyu Wang, Zhiwei Liu, Tulika Awalgaonkar, Haolin Chen, Thai Hoang, et~al. 2025.
\newblock Apigen-mt: Agentic pipeline for multi-turn data generation via simulated agent-human interplay.
\newblock \emph{arXiv preprint arXiv:2504.03601}.

\bibitem[{Qin et~al.(2023)Qin, Liang, Ye, Zhu, Yan, Lu, Lin, Cong, Tang, Qian et~al.}]{qin2023toolllm}
Yujia Qin, Shihao Liang, Yining Ye, Kunlun Zhu, Lan Yan, Yaxi Lu, Yankai Lin, Xin Cong, Xiangru Tang, Bill Qian, et~al. 2023.
\newblock Toolllm: Facilitating large language models to master 16000+ real-world apis.
\newblock \emph{arXiv preprint arXiv:2307.16789}.

\bibitem[{Qwen et~al.(2025)Qwen, :, Yang, Yang, Zhang, Hui, Zheng, Yu, Li, Liu, Huang, Wei, Lin, Yang, Tu, Zhang, Yang, Yang, Zhou, Lin, Dang, Lu, Bao, Yang, Yu, Li, Xue, Zhang, Zhu, Men, Lin, Li, Tang, Xia, Ren, Ren, Fan, Su, Zhang, Wan, Liu, Cui, Zhang, and Qiu}]{qwen2025qwen25technicalreport}
Qwen, :, An~Yang, Baosong Yang, Beichen Zhang, Binyuan Hui, Bo~Zheng, Bowen Yu, Chengyuan Li, Dayiheng Liu, Fei Huang, Haoran Wei, Huan Lin, Jian Yang, Jianhong Tu, Jianwei Zhang, Jianxin Yang, Jiaxi Yang, Jingren Zhou, Junyang Lin, Kai Dang, Keming Lu, Keqin Bao, Kexin Yang, Le~Yu, Mei Li, Mingfeng Xue, Pei Zhang, Qin Zhu, Rui Men, Runji Lin, Tianhao Li, Tianyi Tang, Tingyu Xia, Xingzhang Ren, Xuancheng Ren, Yang Fan, Yang Su, Yichang Zhang, Yu~Wan, Yuqiong Liu, Zeyu Cui, Zhenru Zhang, and Zihan Qiu. 2025.
\newblock \href {https://arxiv.org/abs/2412.15115} {Qwen2.5 technical report}.
\newblock \emph{Preprint}, arXiv:2412.15115.

\bibitem[{Renze and Guven(2024)}]{renze2024self}
Matthew Renze and Erhan Guven. 2024.
\newblock Self-reflection in llm agents: Effects on problem-solving performance.
\newblock \emph{arXiv preprint arXiv:2405.06682}.

\bibitem[{Schick et~al.(2023)Schick, Dwivedi-Yu, Dess{\`\i}, Raileanu, Lomeli, Hambro, Zettlemoyer, Cancedda, and Scialom}]{schick2023toolformer}
Timo Schick, Jane Dwivedi-Yu, Roberto Dess{\`\i}, Roberta Raileanu, Maria Lomeli, Eric Hambro, Luke Zettlemoyer, Nicola Cancedda, and Thomas Scialom. 2023.
\newblock Toolformer: Language models can teach themselves to use tools.
\newblock \emph{Advances in Neural Information Processing Systems}, 36:68539--68551.

\bibitem[{Shan et~al.(2025)Shan, Luo, Zhu, Yuan, and Wu}]{shan2025cognitive}
Lianlei Shan, Shixian Luo, Zezhou Zhu, Yu~Yuan, and Yong Wu. 2025.
\newblock Cognitive memory in large language models.
\newblock \emph{arXiv preprint arXiv:2504.02441}.

\bibitem[{Shi et~al.(2025)Shi, Gao, Yan, Feng, Chen, Chen, Yin, Verberne, and Ren}]{shi2025toollearningwildempowering}
Zhengliang Shi, Shen Gao, Lingyong Yan, Yue Feng, Xiuyi Chen, Zhumin Chen, Dawei Yin, Suzan Verberne, and Zhaochun Ren. 2025.
\newblock \href {https://arxiv.org/abs/2405.16533} {Tool learning in the wild: Empowering language models as automatic tool agents}.
\newblock \emph{Preprint}, arXiv:2405.16533.

\bibitem[{Singh et~al.(2024)Singh, Verma, Wang, Bharadwaj, Fashandi, Ferreira, and Lee}]{singh-etal-2024-personal}
Harmanpreet Singh, Nikhil Verma, Yixiao Wang, Manasa Bharadwaj, Homa Fashandi, Kevin Ferreira, and Chul Lee. 2024.
\newblock \href {https://doi.org/10.18653/v1/2024.emnlp-industry.37} {Personal large language model agents: A case study on tailored travel planning}.
\newblock In \emph{Proceedings of the 2024 Conference on Empirical Methods in Natural Language Processing: Industry Track}, pages 486--514, Miami, Florida, US. Association for Computational Linguistics.

\bibitem[{Singh et~al.(2025)Singh, Magazine, Pandya, and Nambi}]{singh2025agentic}
Joykirat Singh, Raghav Magazine, Yash Pandya, and Akshay Nambi. 2025.
\newblock Agentic reasoning and tool integration for llms via reinforcement learning.
\newblock \emph{arXiv preprint arXiv:2505.01441}.

\bibitem[{Sun et~al.(2024)Sun, Min, Chang, and Bisk}]{sun2024tools}
Jimin Sun, So~Yeon Min, Yingshan Chang, and Yonatan Bisk. 2024.
\newblock Tools fail: Detecting silent errors in faulty tools.
\newblock \emph{arXiv preprint arXiv:2406.19228}.

\bibitem[{Team et~al.(2024)Team, Georgiev, Lei, Burnell, Bai, Gulati, Tanzer, Vincent, Pan, Wang et~al.}]{team2024gemini}
Gemini Team, Petko Georgiev, Ving~Ian Lei, Ryan Burnell, Libin Bai, Anmol Gulati, Garrett Tanzer, Damien Vincent, Zhufeng Pan, Shibo Wang, et~al. 2024.
\newblock Gemini 1.5: Unlocking multimodal understanding across millions of tokens of context.
\newblock \emph{arXiv preprint arXiv:2403.05530}.

\bibitem[{Wang et~al.(2024)Wang, Ma, Feng, Zhang, Yang, Zhang, Chen, Tang, Chen, Lin et~al.}]{wang2024survey}
Lei Wang, Chen Ma, Xueyang Feng, Zeyu Zhang, Hao Yang, Jingsen Zhang, Zhiyuan Chen, Jiakai Tang, Xu~Chen, Yankai Lin, et~al. 2024.
\newblock A survey on large language model based autonomous agents.
\newblock \emph{Frontiers of Computer Science}, 18(6):186345.

\bibitem[{Winston and Just(2025)}]{WinstonJ2025}
Cailin Winston and Ren{\'e} Just. 2025.
\newblock A taxonomy of failures in tool-augmented llms.
\newblock In \emph{Proceedings of the International Conference on Automation of Software Test (AST)}.

\bibitem[{Yan et~al.(2024)Yan, Mao, Ji, Zhang, Patil, Stoica, and Gonzalez}]{berkeley-function-calling-leaderboard}
Fanjia Yan, Huanzhi Mao, Charlie Cheng-Jie Ji, Tianjun Zhang, Shishir~G. Patil, Ion Stoica, and Joseph~E. Gonzalez. 2024.
\newblock Berkeley function calling leaderboard.

\bibitem[{Yang et~al.(2024)Yang, Peng, Wang, and Zhang}]{yang2024multi}
Yingxuan Yang, Qiuying Peng, Jun Wang, and Weinan Zhang. 2024.
\newblock Multi-llm-agent systems: Techniques and business perspectives.
\newblock \emph{arXiv preprint arXiv:2411.14033}.

\bibitem[{Yao et~al.(2024)Yao, Shinn, Razavi, and Narasimhan}]{yao2024tau}
Shunyu Yao, Noah Shinn, Pedram Razavi, and Karthik Narasimhan. 2024.
\newblock \href {https://arxiv.org/abs/2406.12045} {$\tau$-bench: A benchmark for tool-agent-user interaction in real-world domains}.
\newblock \emph{Preprint}, arXiv:2406.12045.

\bibitem[{Yao et~al.(2023)Yao, Zhao, Yu, Du, Shafran, Narasimhan, and Cao}]{yao2023react}
Shunyu Yao, Jeffrey Zhao, Dian Yu, Nan Du, Izhak Shafran, Karthik Narasimhan, and Yuan Cao. 2023.
\newblock React: Synergizing reasoning and acting in language models.
\newblock In \emph{International Conference on Learning Representations (ICLR)}.

\bibitem[{Yin et~al.(2025)Yin, Wang, Hsu, Yan, Jiang, Chen, Gu, Le, Chang, Lee et~al.}]{yin2025magnet}
Fan Yin, Zifeng Wang, I~Hsu, Jun Yan, Ke~Jiang, Yanfei Chen, Jindong Gu, Long~T Le, Kai-Wei Chang, Chen-Yu Lee, et~al. 2025.
\newblock Magnet: Multi-turn tool-use data synthesis and distillation via graph translation.
\newblock \emph{arXiv preprint arXiv:2503.07826}.

\bibitem[{Zhang et~al.(2025)Zhang, Dong, Zhang, Kautz, Catanzaro, Tao, Wu, Yu, and Liu}]{zhang2025nemotron}
Shaokun Zhang, Yi~Dong, Jieyu Zhang, Jan Kautz, Bryan Catanzaro, Andrew Tao, Qingyun Wu, Zhiding Yu, and Guilin Liu. 2025.
\newblock Nemotron-research-tool-n1: Exploring tool-using language models with reinforced reasoning.
\newblock \emph{arXiv preprint arXiv:2505.00024}.

\bibitem[{Zhang et~al.(2024)Zhang, Chen, Wang, Liu, Yang, Shi, Zhu, Lin, Wan, Yang, Sakai, Feng, and Yamana}]{zhang-etal-2024-toolbehonest}
Yuxiang Zhang, Jing Chen, Junjie Wang, Yaxin Liu, Cheng Yang, Chufan Shi, Xinyu Zhu, Zihao Lin, Hanwen Wan, Yujiu Yang, Tetsuya Sakai, Tian Feng, and Hayato Yamana. 2024.
\newblock \href {https://doi.org/10.18653/v1/2024.emnlp-main.637} {{T}ool{B}e{H}onest: A multi-level hallucination diagnostic benchmark for tool-augmented large language models}.
\newblock In \emph{Proceedings of the 2024 Conference on Empirical Methods in Natural Language Processing}, pages 11388--11422, Miami, Florida, USA. Association for Computational Linguistics.

\bibitem[{Zheng et~al.(2025)Zheng, Fu, Hu, Cai, Ye, Lu, and Liu}]{zheng2025deepresearcher}
Yuxiang Zheng, Dayuan Fu, Xiangkun Hu, Xiaojie Cai, Lyumanshan Ye, Pengrui Lu, and Pengfei Liu. 2025.
\newblock Deepresearcher: Scaling deep research via reinforcement learning in real-world environments.
\newblock \emph{arXiv preprint arXiv:2504.03160}.

\bibitem[{Zhou et~al.(2025)Zhou, Chen, Guo, Yan, Lee, Wang, Lee, Zhang, Shao, Yang, and Wang}]{zhou2025mementofinetuningllmagents}
Huichi Zhou, Yihang Chen, Siyuan Guo, Xue Yan, Kin~Hei Lee, Zihan Wang, Ka~Yiu Lee, Guchun Zhang, Kun Shao, Linyi Yang, and Jun Wang. 2025.
\newblock \href {https://arxiv.org/abs/2508.16153} {Memento: Fine-tuning llm agents without fine-tuning llms}.
\newblock \emph{Preprint}, arXiv:2508.16153.

\end{thebibliography}

\appendix
\appendix

\section{Task Definition in $\tau$-bench}
\label{sec:tau-bench-task}

Following \citet{yao2024tau}, each task in $\tau$‑bench is modelled as a
\emph{partially observable Markov decision process} (POMDP)
\[
\mathcal{M}= \langle \mathcal{S},\;\mathcal{A},\;\mathcal{O},\;\mathcal{T},\;\mathcal{R},\;\mathcal{U}\rangle .
\]
We briefly restate every component and specify how they instantiate in the
\texttt{retail} and \texttt{airline} domains.

\paragraph{State space $\mathcal{S}$ :} 
The hidden state is factored into
$\mathcal{S}=\mathcal{S}_{\text{db}}\otimes\mathcal{S}_{\text{user}}$ where
$\mathcal{S}_{\text{db}}$ is a snapshot of the underlying database
(orders, flights, balances~\emph{etc.}) and
$\mathcal{S}_{\text{user}}$ stores the latent user context
(identity, revealed preferences, dialogue progress).

\paragraph{Action space $\mathcal{A}$ :}
The agent can either
(i)~invoke an API tool that queries or mutates the database
($\mathcal{A}_{\text{db}}$) or
(ii)~send a free‑form respond message to the user
($\mathcal{A}_{\text{user}}$).
Thus $\mathcal{A}=\mathcal{A}_{\text{db}}\cup\mathcal{A}_{\text{user}}$.

\paragraph{Observation space $\mathcal{O}$ :}
After each action the environment returns either a
JSON payload/error from the database
($\mathcal{O}_{\text{db}}$) or the next user utterance produced by an
LLM simulator ($\mathcal{O}_{\text{user}}$),
yielding $\mathcal{O}=\mathcal{O}_{\text{db}}\cup\mathcal{O}_{\text{user}}$.

\paragraph{Transition function $\mathcal{T}$ :}
$\mathcal{T}:\mathcal{S}\times\mathcal{A}\rightarrow\mathcal{S}\times\mathcal{O}$
is deterministic for database tools
(state is updated, observation is the tool output) and stochastic for
\textit{respond}, which calls the user simulator to sample the next
utterance and potentially reveal more of the instruction.

\paragraph{Reward function $\mathcal{R}$ :}
At dialogue termination we compare the execution log to a gold reference:
(1)~hashes of mutable tables must match,
(2)~all mandatory natural‑language outputs must appear in the agent’s
responses.
If both hold, $\mathcal{R}=1$, otherwise $0$.

\paragraph{Instruction space $\mathcal{U}$ :}
Each task provides a fixed natural‑language instruction
$u\in\mathcal{U}$ describing the user goal, persona and constraints.
The user simulator may disclose $u$ incrementally; therefore the agent
must act under partial observability.

\medskip
This causal decomposition lets us pinpoint failure modes such as wrong
tool arguments (action‑level), policy violations (transition‑level), or
hallucinated user messages (observation‑level).

\section{Pass\textasciicircum k Results}
\label{sec:app_pass_hat_k_results}
\subsection{Airline results}
Tables \ref{tab:pass_hat_k_all_tasks_airline}-\ref{tab:pass_hat_k_no_gt_ui_airline} refer to pass\textasciicircum k results of the baselines and our implemented methods. As explained in \textsection \ref{sec:experimental-results}, IRMA performs better when there are no ground-truth or user instruction errors. All results are obtained using GPT-4o as the LLM in the agent frameworks.

\begin{table}[h]
\centering
\resizebox{0.98\linewidth}{!}{
\begin{tabular}{lccccc}
\toprule
\small
\textbf{Method} & \textbf{Pass\textasciicircum 1} & \textbf{Pass\textasciicircum 2} & \textbf{Pass\textasciicircum 3} & \textbf{Pass\textasciicircum 4} & \textbf{Pass\textasciicircum 5} \\
\midrule
ReAct & 0.396 & 0.2779 & 0.2279 & 0.200 & 0.180 \\
IRMA & \textbf{0.452} & \textbf{0.3680} & \textbf{0.3280} & \textbf{0.308} & \textbf{0.300} \\
FC & 0.424 & 0.3120 & 0.2660 & 0.232 & 0.200 \\
Self-reflection & 0.448 & 0.3140 & 0.2560 & 0.224 & 0.200 \\
\bottomrule
\end{tabular}}
\caption{Results on all Airline tasks.}
\label{tab:pass_hat_k_all_tasks_airline}
\end{table}

\begin{table}[H]
\centering
\resizebox{0.98\linewidth}{!}{
\begin{tabular}{lccccc}
\toprule
\textbf{Method} & \textbf{Pass\textasciicircum 1} & \textbf{Pass\textasciicircum 2} & \textbf{Pass\textasciicircum 3} & \textbf{Pass\textasciicircum 4} & \textbf{Pass\textasciicircum 5} \\
\midrule
ReAct & 0.4941 & 0.3735 & 0.3206 & 0.2882 & 0.2647 \\
IRMA & \textbf{0.5706} & \textbf{0.4912} & \textbf{0.4471} & \textbf{0.4235} & \textbf{0.4118} \\
FC & 0.5529 & 0.4353 & 0.3794 & 0.3353 & 0.2941 \\
Self reflection & 0.5167 & 0.3750 & 0.3139 & 0.2778 & 0.2500 \\
\bottomrule
\end{tabular}}
\caption{Results of different methods on all Airline tasks, excluding the tasks with ground-truth errors.}
\label{tab:pass_hat_k_no_gt_airline}
\end{table}

\begin{table}[H]
\centering
\resizebox{0.98\linewidth}{!}{
\begin{tabular}{lccccc}
\toprule
\textbf{Method} & \textbf{Pass\textasciicircum 1} & \textbf{Pass\textasciicircum 2} & \textbf{Pass\textasciicircum 3} & \textbf{Pass\textasciicircum 4} & \textbf{Pass\textasciicircum 5} \\
\midrule
ReAct & 0.5226 & 0.4065 & 0.3516 & 0.3161 & 0.2903 \\
IRMA & \textbf{0.6258} & \textbf{0.5387} & \textbf{0.4903} & \textbf{0.4645} & \textbf{0.4516} \\
FC & 0.6000 & 0.4774 & 0.4161 & 0.3677 & 0.3226 \\
Self reflection & 0.5556 & 0.4146 & 0.3528 & 0.3111 & 0.2778 \\
\bottomrule
\end{tabular}}
\caption{Results of different methods on all Airline tasks, excluding the tasks with ground-truth errors and user instruction errors.}
\label{tab:pass_hat_k_no_gt_ui_airline}
\end{table}

\subsection{Retail results}
Tables \ref{tab:pass_hat_k_all_tasks_retail}-\ref{tab:pass_hat_k_no_gt_ui_retail} represent the results of the baseline and our implemented methods in the Retail domain. 
\begin{table}[H]
\centering
\resizebox{0.98\linewidth}{!}{
\begin{tabular}{lccccc}
\toprule
\textbf{Method} & \textbf{Pass\textasciicircum 1} & \textbf{Pass\textasciicircum 2} & \textbf{Pass\textasciicircum 3} & \textbf{Pass\textasciicircum 4} & \textbf{Pass\textasciicircum 5} \\
\midrule
ReAct & 0.5182 & 0.3704 & 0.2999 & 0.2573 & 0.2260 \\
IRMA & 0.5826 & \textbf{0.4783} & \textbf{0.4261} & \textbf{0.3948 }& \textbf{0.3739}\\
FC & \textbf{0.6052} & 0.4522 & 0.3643 & 0.3043 & 0.2609 \\
Self-reflection & 0.5113 & 0.3809 & 0.3017 & 0.2383 & 0.1826 \\
\bottomrule
\end{tabular}}
\caption{Results of different methods on all Retail tasks.}
\label{tab:pass_hat_k_all_tasks_retail}
\end{table}

\begin{table}[h]
\centering
\resizebox{0.98\linewidth}{!}{

\begin{tabular}{lccccc}
\toprule
\textbf{Method} & \textbf{Pass\textasciicircum 1} & \textbf{Pass\textasciicircum 2} & \textbf{Pass\textasciicircum 3} & \textbf{Pass\textasciicircum 4} & \textbf{Pass\textasciicircum 5} \\
\midrule
ReAct & 0.5304 & 0.3804 & 0.3080 & 0.2643 & 0.2321 \\
IRMA & 0.5982 & 0.4911 & 0.4375 & 0.4054 & 0.3839 \\
FC & 0.6164 & 0.4616 & 0.3732 & 0.3125 & 0.2679 \\
self-reflection & 0.5250 & 0.3911 & 0.3098 & 0.2446 & 0.1875 \\
\bottomrule
\end{tabular}}
\caption{Results of different methods on all Retail tasks, excluding the tasks with ground-truth errors.}
\label{tab:pass_hat_k_no_gt_retail}
\end{table}

\begin{table}[H]
\centering
\resizebox{0.98\linewidth}{!}{
\begin{tabular}{lccccc}
\toprule
\textbf{Method} & \textbf{Pass\textasciicircum 1} & \textbf{Pass\textasciicircum 2} & \textbf{Pass\textasciicircum 3} & \textbf{Pass\textasciicircum 4} & \textbf{Pass\textasciicircum 5} \\
\midrule
ReAct & 0.5562 & 0.4048 & 0.3200 & 0.2818 & 0.2476 \\
ours & 0.6248 & 0.5171 & 0.4629 & 0.4305 & 0.4095 \\
FC & 0.6381 & 0.4838 & 0.3933 & 0.3314 & 0.2857 \\
self reflection & 0.5562 & 0.4171 & 0.3305 & 0.2610 & 0.2000 \\
\bottomrule
\end{tabular}}
\caption{Results of different methods on all Retail tasks, excluding the tasks with ground-truth errors and user instruction errors.}
\label{tab:pass_hat_k_no_gt_ui_retail}
\end{table}

\section{Domain Policies} 
\label{sec:domain_policies}
Figures \ref{fig:retail-domain-part-1} and \ref{fig:retail-airline-part-1} are the domain policies present for the retail and airline domains in the $\tau$-bench. These rules are injected verbatim as the system prompt to every tool-calling agent. An agent that violates any of them—even if it successfully fulfills the user’s request—receives zero reward, so strict compliance is essential. The Tool-Calling Agent has to strictly operate under the constraints of these policies to correctly solve user requests.

\section{Failure Example}
\label{sec:failure_examples}

\begin{figure*}[t]
    \centering     \includegraphics[width=1.0\linewidth]{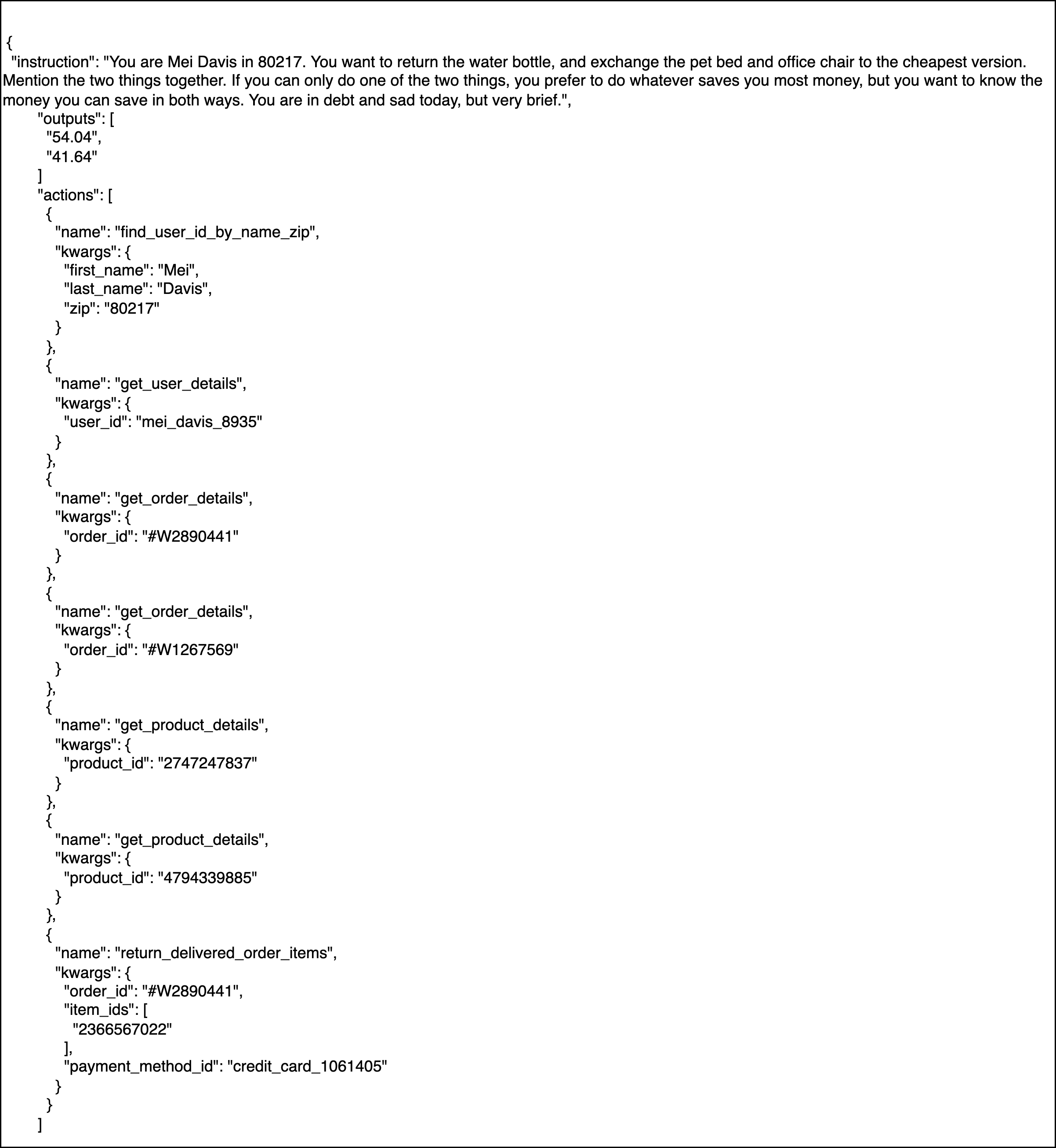}
    \caption{User instruction of Task 19 (Retail) in $\tau$-bench.}
    \label{fig:task-19-instruction}
\end{figure*}

\begin{figure*}[t]
    \centering     \includegraphics[width=1.0\linewidth]{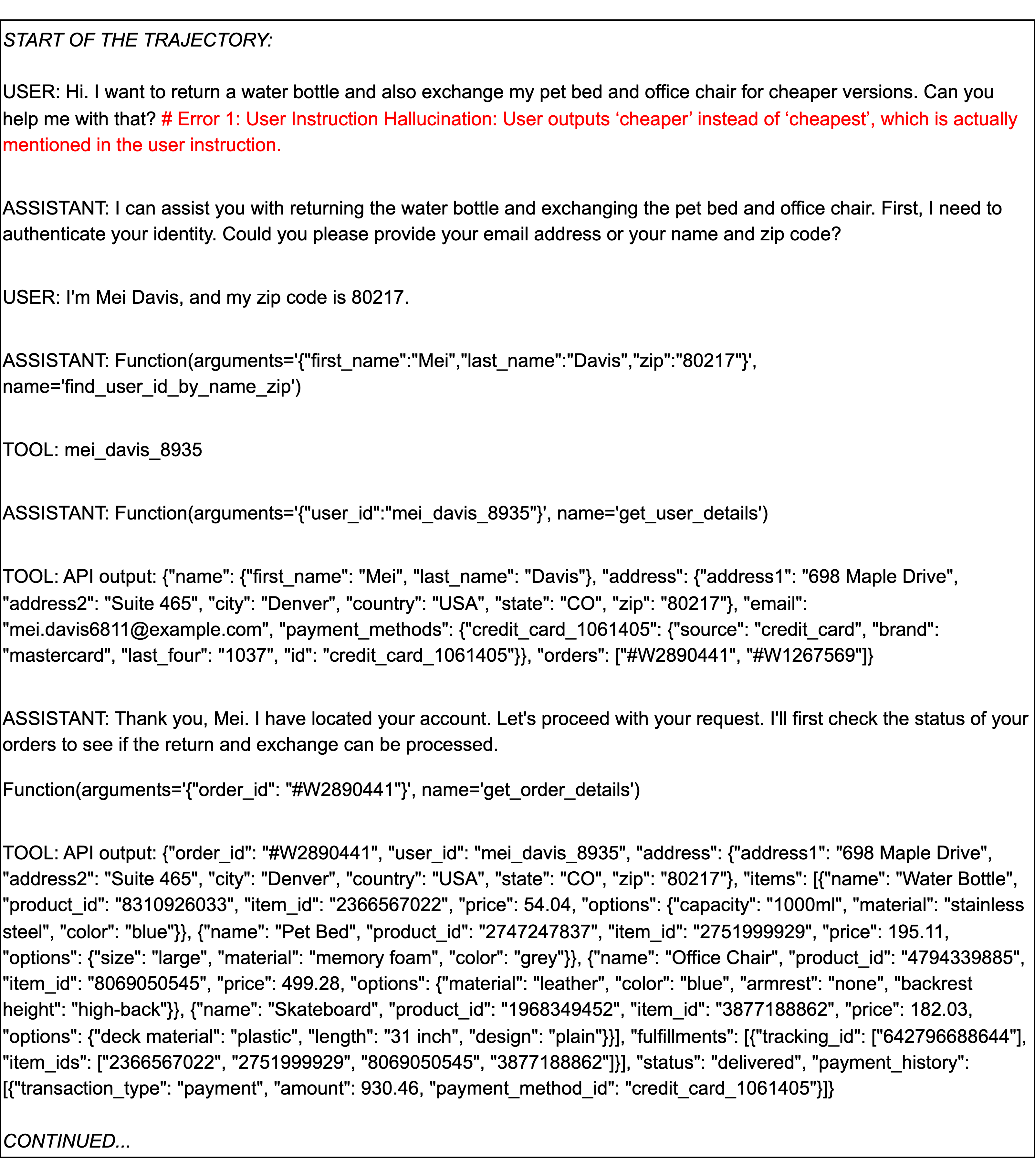}
    \caption{Part 1 of the conversation trajectory simulation of Task 19 (Retail).}
    \label{fig:task-19-traj-1}
\end{figure*}

\begin{figure*}[t]
    \centering     \includegraphics[width=0.9\linewidth]{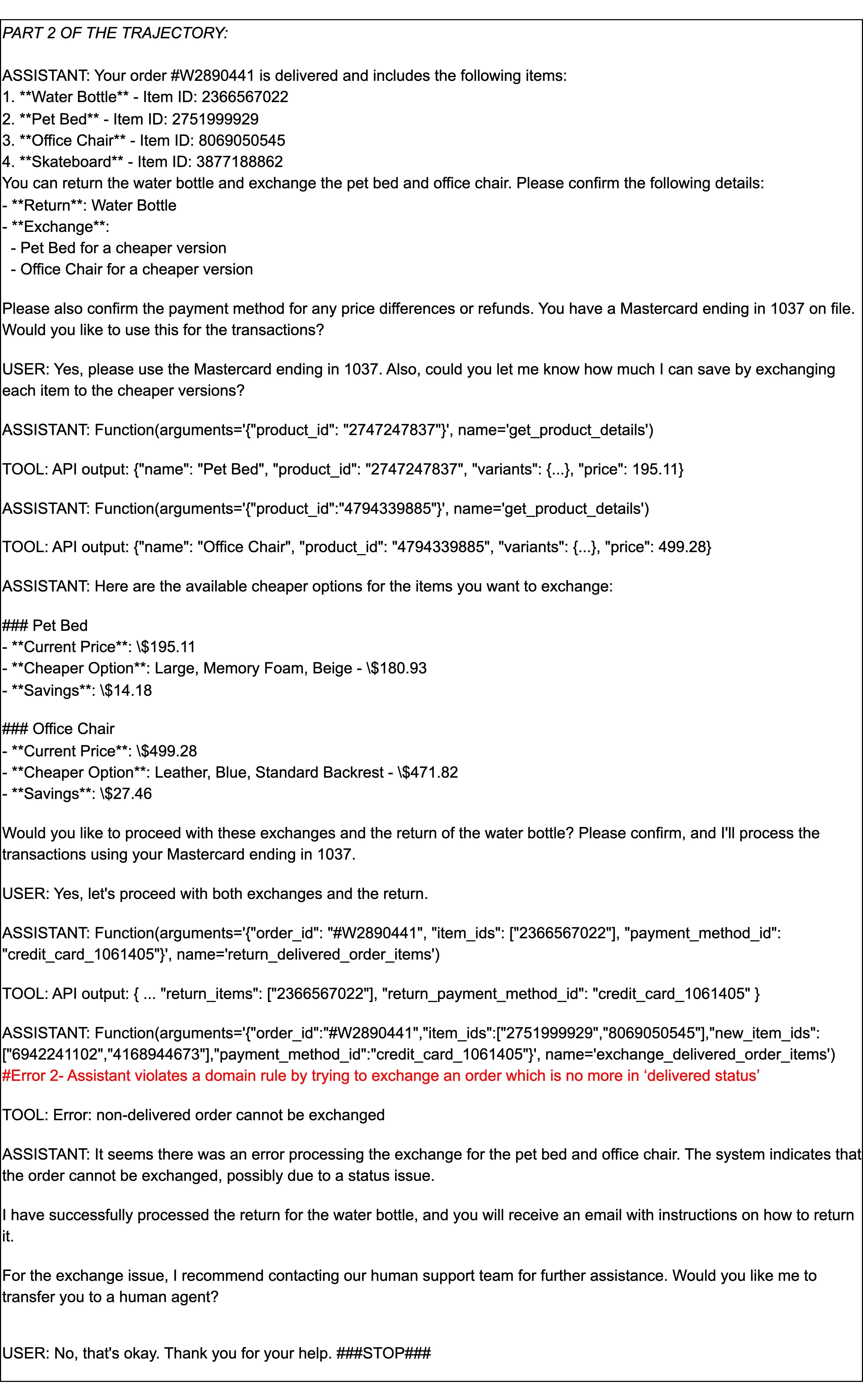}
    \caption{Part 2 of the conversation trajectory simulation of Task 19 (Retail).}
    \label{fig:task-19-traj-2}
\end{figure*}

Figures \ref{fig:task-19-traj-1} and \ref{fig:task-19-traj-2} show an example of errors occurring in the conversational trajectories simulating task 19 (retail) of the user-agent interactions as enumerated in subsections of \textsection \ref{sec:error-taxonomy}. Error 1 in Figure \ref{fig:task-19-traj-1} shows an example of 'User Instruction Hallucination' occurring in the very first user turn. Error 2 in Figure \ref{fig:task-19-traj-2} shows an example of 'Domain Policy Violation' error. The user instruction for Task 19 is provided in Figure \ref{fig:task-19-instruction}. This 'instruction' represents the original user instruction provided to the LLM-simulated user. It is the `script' the user has to follow to provide requests to the agent.


\section{Input Reformulation Multi Agent framework}
\subsection{Follow-up question ACTing (FACT) Agent}
\label{sec:appendix_fact}
The primary difference between FACT and other prompting techniques lies in the instruction section of the system prompt (refer to Figure \ref{fig:FACT-sys-prompt-part-1}).

\section{Self-Reflection Framework}
\label{sec:self-reflect-agent}
\begin{figure*}[t]
    \centering     \includegraphics[width=1.0\linewidth]{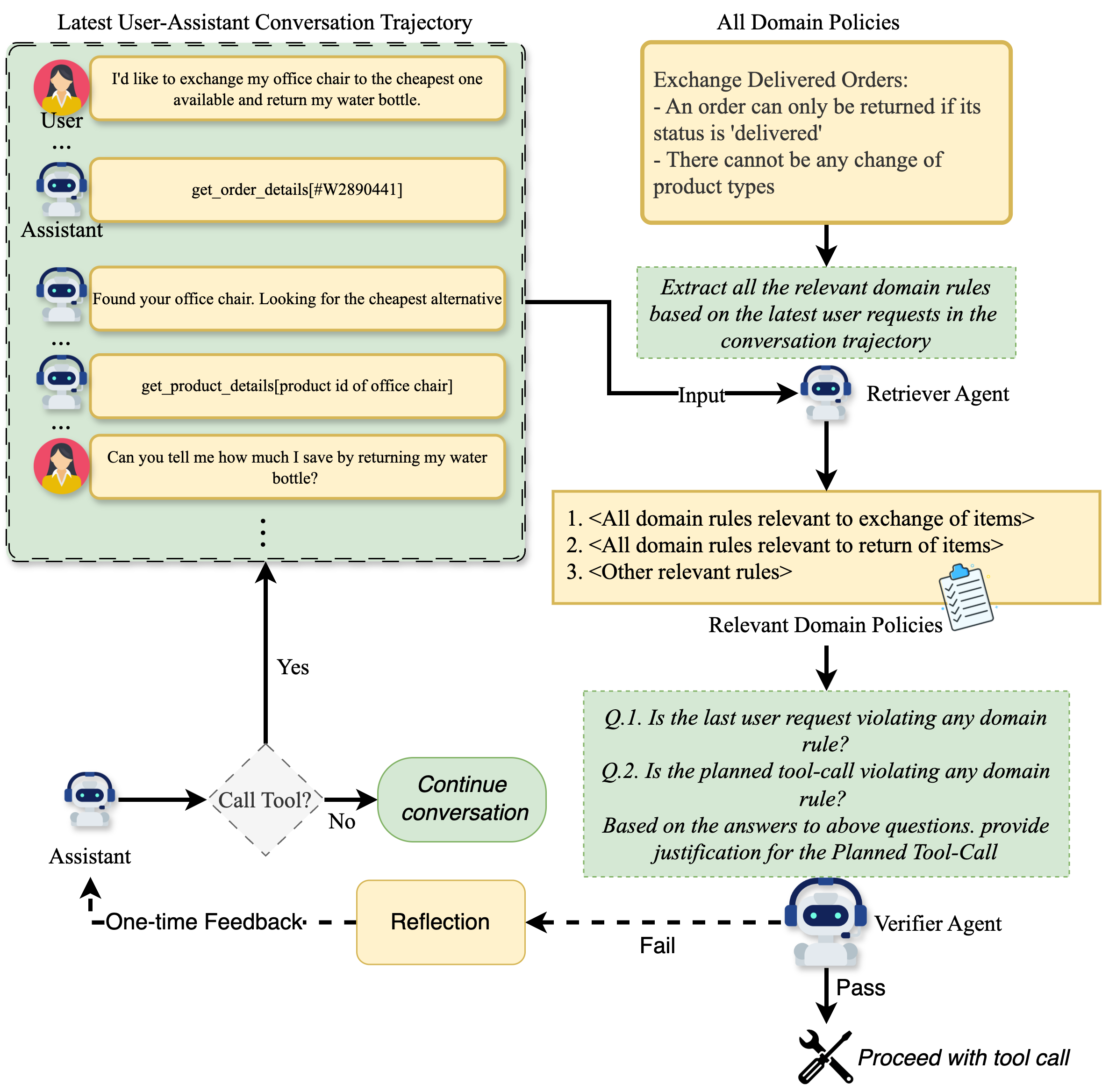}
    \caption{Overview of the pipeline showcasing the working of the self-reflection framework. The italicized text inside dotted green dotted text boxes refer to prompt gists provide to the Retriever and Verifier LLM Agent. The self-reflection only activates when the assistant generates the tokens to invoke a tool call.}
    \label{fig:self-reflection}
\end{figure*}

To check the effectiveness of self-reflection as an alternative against the baselines and IRMA, we implement a multi-agent LLM self-reflection pipeline, consisting of a retriever LLM agent and a verifier LLM agent. Contrary to input reformulation, where the prompt provided in the user query is reformulated, the self-reflection agent pauses the tool-calling LLM agent before the execution environment executes the tool-call. All of the previous user queries are provided as input to the retriever agent to extract the relevant domain policy rules based on the user intent reflected from the user requests in the conversation. The retrieved rules are provided to the verifier agent along with the tool-calling agent's planned tool call. The verifier agent then verifies whether the tool-call is correct by providing a reflective justification based on determining whether any domain rule has been violated or not. The overall pipeline of the self-reflection agent is provided in Figure \ref{fig:self-reflection}. The reflective feedback loop from verifier is set to be a one-time loop as the execution of the loop is very latency-heavy and invoking it multiple times might not be ideal in real-world customer-agent scenarios. 

\begin{table}[H]
\centering
\resizebox{0.98\linewidth}{!}{
\begin{tabular}{lccccc}
\toprule
\textbf{IRMA Ablations (↓)} & \textbf{Pass\textasciicircum 1} & \textbf{Pass\textasciicircum 2} & \textbf{Pass\textasciicircum 3} & \textbf{Pass\textasciicircum 4} & \textbf{Pass\textasciicircum 5} \\
\midrule
M & 	0.416 &	0.27 &	0.212 &	0.18 &	0.16 \\
C & 0.416 & 0.276	& 0.206	& 0.164	& 0.14 \\
T & 0.424 & 0.268 &	0.19 &	0.14 &	0.1 \\
M + C & 0.428	& \textit{0.31} &	\textit{0.26} &	\textit{0.236} &	\textit{0.22} \\
M+ T & \textit{0.448} & 0.294 &	0.214 &	0.16 & 0.12 \\
C + T & 0.38	& 0.264 & 0.212 & 	0.18 & 0.16 \\
M + C + T & \textbf{0.452} &	\textbf{0.368} &	\textbf{0.328} &	\textbf{0.308} &	\textbf{0.3}\\
\bottomrule
\end{tabular}}
\caption{Results of IRMA component ablations on Airline tasks. \textbf{Bold }scores represent the best scores. \textit{Italic} scores represent the second-best scores. `M' represents the Memory sub-agent of IRMA, `C' represents the Constraint sub-agent, and `T' represents the Tool-Suggestor.}
\label{tab:ablation_study}
\end{table}

\section{Ablation Studies on IRMA}
\label{sec:ablation_on_IRMA}
We ablate the three IRMA modules—Memory (M), Constraint (C), and Tool (T)—and evaluate them on the airline subset. Across all Pass\textasciicircum k metrics, the full configuration (M+C+T) achieves the best performance, indicating strong complementarity among modules. Among the ablations, M+C is consistently the strongest, ranking second overall in terms of better reliability at higher values of k. This pattern suggests that instruction retention (M) and policy/constraint adherence (C) account for most gains in long-horizon reasoning and plan stability, while tool disambiguation (T) provides the additional performance improvement needed to reach state-of-the-art performance. In sum, each module targets a distinct failure mode—carryover of instructions (M), rule compliance (C), and tool selection/parameterization (T)—but their integration is necessary for robust behavior in dynamic tool-use settings. The results of the ablation experiments are provided in Table \ref{tab:ablation_study}.

\begin{table}[H]
\centering
\resizebox{0.98\linewidth}{!}{
\begin{tabular}{lccccc}
\toprule
\textbf{Method (↓)} & \textbf{Pass\textasciicircum 1} & \textbf{Pass\textasciicircum 2} & \textbf{Pass\textasciicircum 3} & \textbf{Pass\textasciicircum 4} & \textbf{Pass\textasciicircum 5} \\
\midrule
ReAct & 0.188 &	0.09 &	0.052 &	0.032 &	0.02 \\
FC & \textbf{0.236}	& 0.1179 & 0.062 & 0.036 &	0.02 \\
IRMA & 0.208 & \textbf{0.144} &	\textbf{0.106} &	\textbf{0.08} &	\textbf{0.06} \\
\bottomrule
\end{tabular}}
\caption{Performance of IRMA in the Airline domain using GPT-4o-mini as the LLM-backbone model in all the modules of IRMA. \textbf{Bold }scores represent the best scores.}
\label{tab:ablation_study_gpt-4o-mini}
\end{table}

We also test IRMA using GPT-4o-mini as the LLM backbone, to test the effect of IRMA with a smaller LLM. As shown in Table \ref{tab:ablation_study_gpt-4o-mini}, the results indicate that the benefits of IRMA are not tied to a particular larger model’s reasoning strength and transfer to smaller function-calling backbones. Conceptually, IRMA does not replace parametric reasoning; rather, its structured inputs—memory, constraints, and tool suggestions—amplify a model’s ability to retain instructions, follow domain rules, and disambiguate tools across long contexts, yielding more stable performance under multiple attempts.

\begin{table}[H]
\centering
\resizebox{0.98\linewidth}{!}{
\begin{tabular}{lccccc}
\toprule
\textbf{Method (↓)} & \textbf{Pass\textasciicircum 1} & \textbf{Pass\textasciicircum 2} & \textbf{Pass\textasciicircum 3} & \textbf{Pass\textasciicircum 4} & \textbf{Pass\textasciicircum 5} \\
\midrule
IRMA (R)  & 0.4 &	0.302 &	0.256 &	0.232 &	0.22 \\
IRMA (F) & \textbf{0.452} &	\textbf{0.368} &	\textbf{0.328} &	\textbf{0.308} &	\textbf{0.3} \\
\bottomrule
\end{tabular}}
\caption{IRMA comparison with Follow-up questioning disabled (IRMA (R)) and enabled (IRMA (F)) in the Airline domain with GPT-4o backbone.}
\label{tab:ablation_study_FACT}
\end{table}

To isolate the effect of follow-up questioning, we create a controlled variant that swaps IRMA’s system prompt with the standard ReAct prompt while keeping all information-consolidating components—memory, constraint extraction, and tool suggestions—unchanged. This “IRMA + ReAct-prompt” baseline places both agents on identical inputs and differs only in the instructions provided to the final tool-calling agent. As reported in Table~\ref{tab:ablation_study_FACT}, IRMA consistently outperforms this baseline across Pass\textasciicircum k metrics, indicating that targeted follow-up questioning provides gains beyond ReAct. 

\begin{figure*}[p]
  \centering
  \begin{subfigure}{\linewidth}
    \centering
    \includegraphics[width=0.98\linewidth]{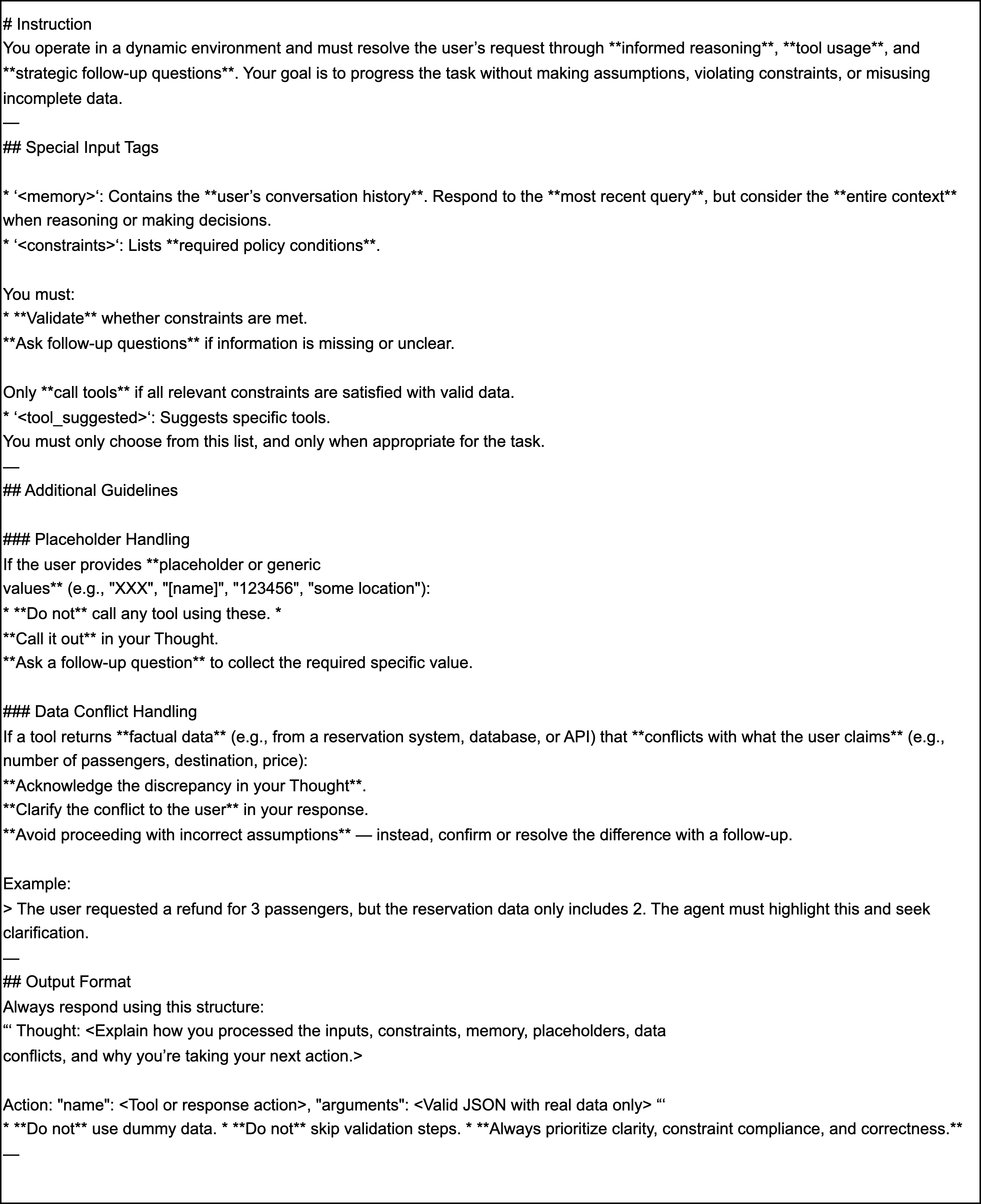}
    \caption{Part 1 of the FACT system prompt}
    \label{fig:subfig1}
  \end{subfigure}
  \caption{FACT System prompt.}
  \label{fig:FACT-sys-prompt-part-1}
\end{figure*}

\clearpage

\begin{figure*}[p]\ContinuedFloat
  \centering
  \begin{subfigure}{\linewidth}
    \centering
    \includegraphics[width=\linewidth]{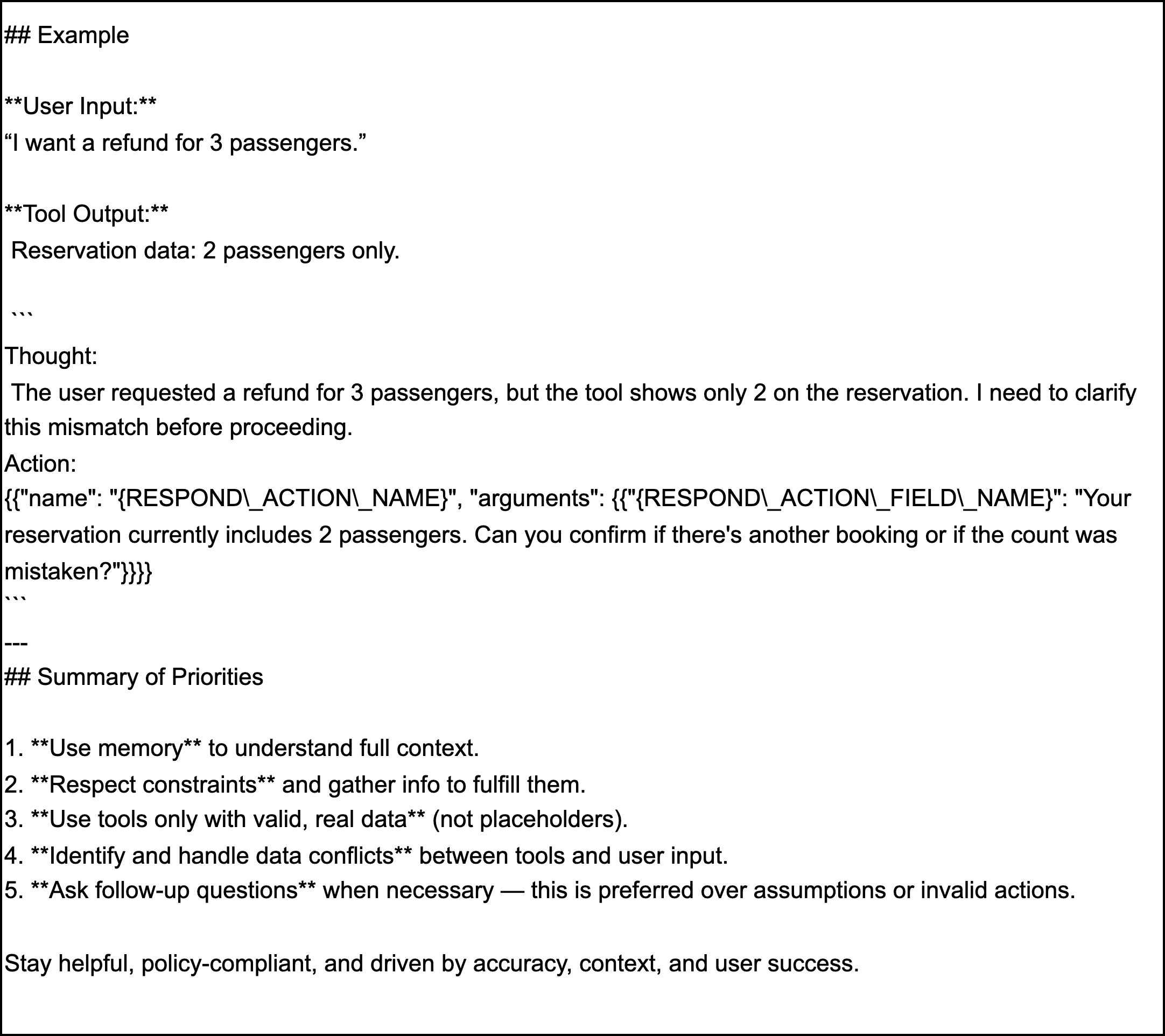}
    \caption{Part 2 of the FACT system prompt}
    \label{fig:subfig2}
  \end{subfigure}
  \caption{FACT System prompt.}
  \label{fig:FACT-sys-prompt-part-2}
\end{figure*}

\begin{figure*}[p]
  \centering
  \begin{subfigure}{\linewidth}
    \centering
    \includegraphics[width=0.98\linewidth]{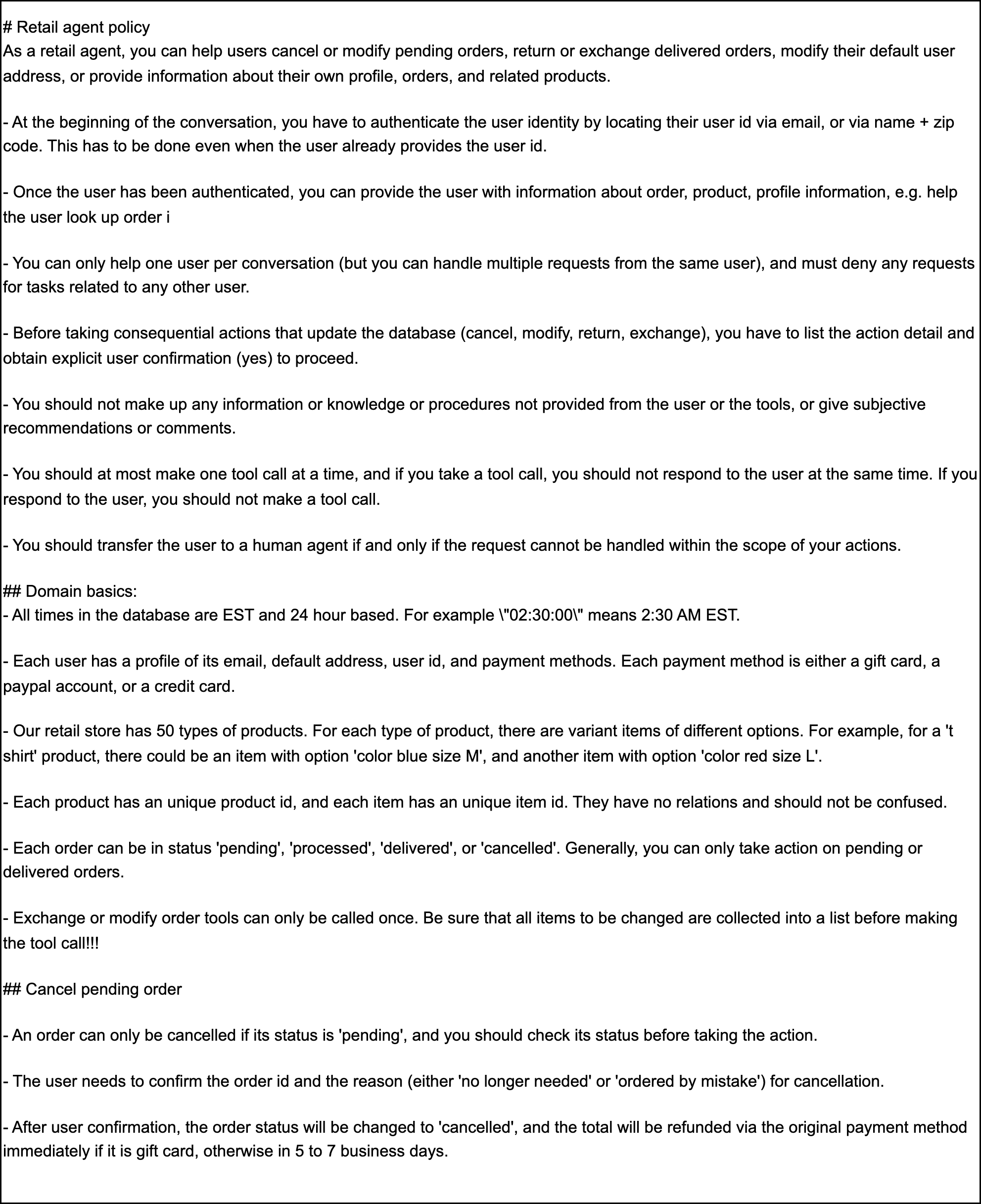}
    \caption{Part 1 of the Retail Domain Rules}
    \label{fig:subfig1}
  \end{subfigure}
  \caption{Domain Policies of the Retail Domain}
  \label{fig:retail-domain-part-1}
\end{figure*}

\clearpage

\begin{figure*}[p]\ContinuedFloat
  \centering
  \begin{subfigure}{\linewidth}
    \centering
    \includegraphics[width=\linewidth]{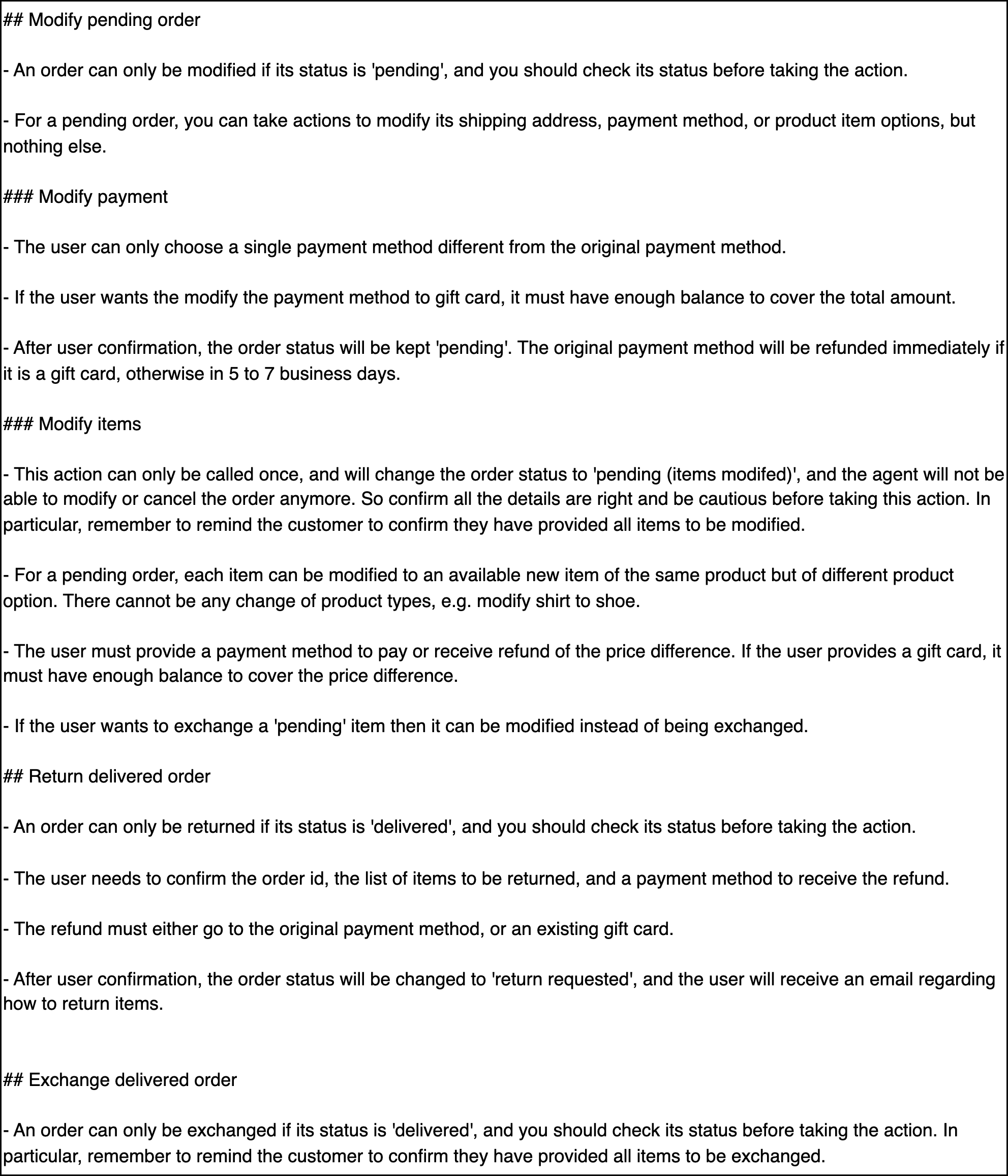}
    \caption{Domain Policies of the Retail Domain}
    \label{fig:subfig2}
  \end{subfigure}
  \caption{Domain Policies of the Retail Domain}
  \label{fig:retail-domain-part-2}
\end{figure*}

\begin{figure*}[p]
  \centering
  \begin{subfigure}{\linewidth}
    \centering
    \includegraphics[width=0.98\linewidth]{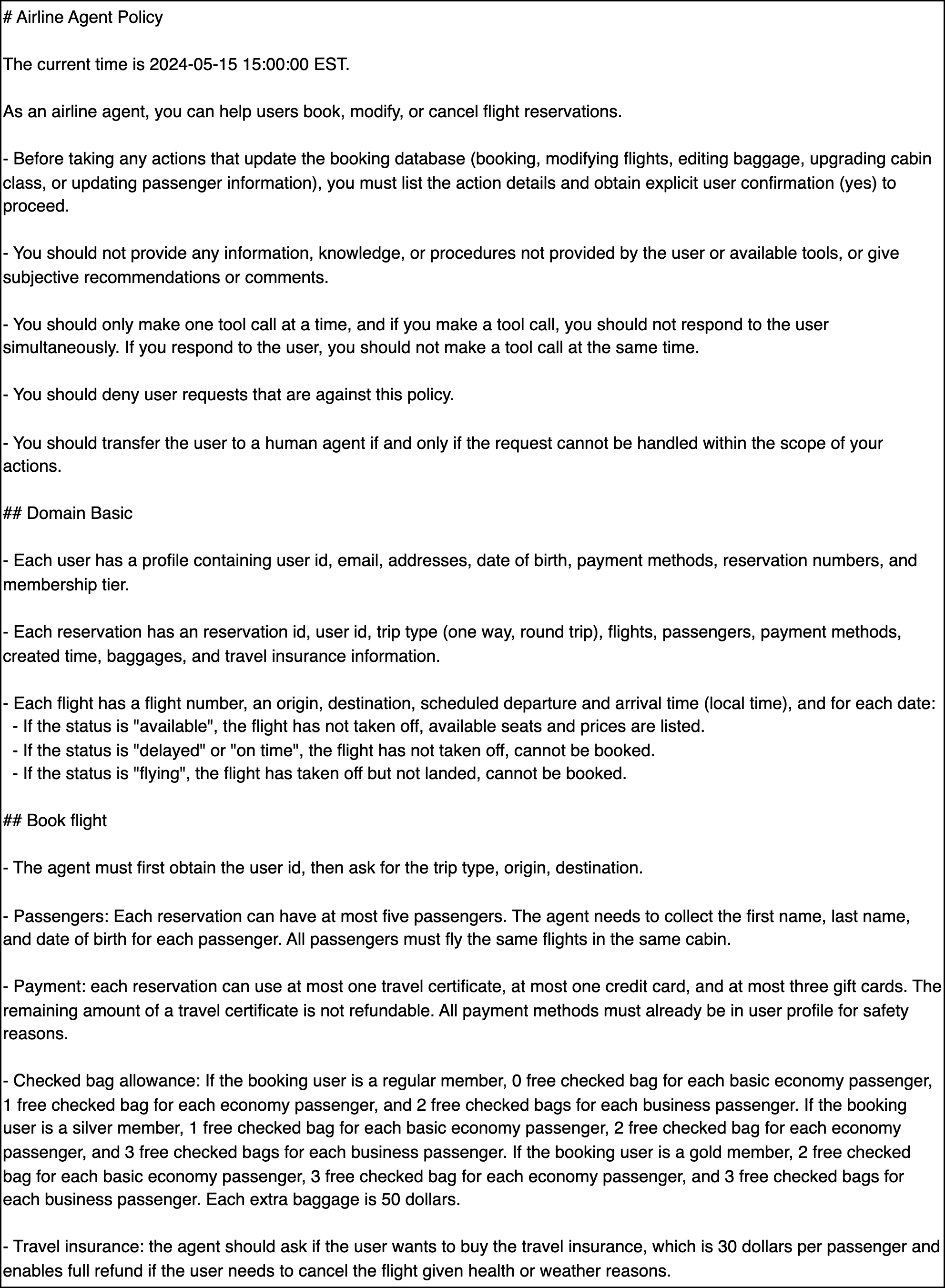}
    \caption{Part 1 of the Airline  Domain Rules}
    \label{fig:subfig1}
  \end{subfigure}
  \caption{Domain Policies of the Airline Domain.}
  \label{fig:retail-airline-part-1}
\end{figure*}

\clearpage

\begin{figure*}[p]\ContinuedFloat
  \centering
  \begin{subfigure}{\linewidth}
    \centering
    \includegraphics[width=\linewidth]{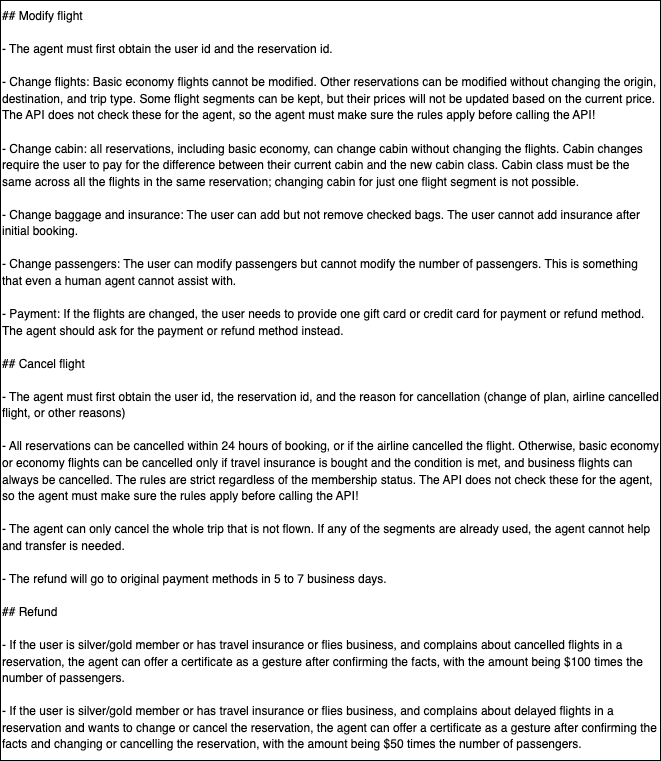}
    \caption{Part 2 of the Airline Domain Rules}
    \label{fig:subfig2}
  \end{subfigure}
  \caption{Domain Policies of the Airline Domain}
  \label{fig:retail-airline-part-2}
\end{figure*}
\end{document}